\documentclass[11pt]{article}

\usepackage{acl}

\usepackage{times}
\usepackage{latexsym}

\usepackage[T1]{fontenc}

\usepackage[utf8]{inputenc}

\usepackage{microtype}
\usepackage{inconsolata}

\usepackage{amsmath}
\usepackage{amssymb}

\usepackage{booktabs}
\usepackage{multirow}
\usepackage{tabularx}
\usepackage{makecell}
\usepackage[table]{xcolor}

\definecolor{HeaderBg}{HTML}{EFEFEF}
\definecolor{GapBg}{HTML}{FCE8E6}
\definecolor{SoftGreen}{HTML}{E6F4EA}
\definecolor{SoftYellow}{HTML}{FFF4CE}

\usepackage{graphicx}
\usepackage{adjustbox}
\usepackage{url}
\usepackage{stfloats}
\usepackage[font=small,skip=3pt]{caption}

\usepackage{enumitem}
\usepackage{placeins}
\usepackage{float}

\usepackage{xspace}

\setlist[itemize]{leftmargin=*, topsep=2pt, itemsep=1pt}
\setlist[enumerate]{leftmargin=*, topsep=2pt, itemsep=1pt}

\newcommand{\method}{VISA\xspace}
\newcommand{\actcheck}{ActCheck\xspace}

\newcommand{\hua}[1]

\title{\textsc{STAGE}: Diagnosing Semantic Transfer at Grounded Execution in Embodied Agents}

\author{
  \textbf{Baosheng Jin} \quad
  \textbf{Yushen Liang} \quad
  \textbf{Hua Shen} \\
  Center for Data Science, NYU Shanghai \\
  Shanghai, China \\
  \texttt{\{bj2410, yl11949, hs3645\}@nyu.edu}
}

\begin{document}
\maketitle

\begin{abstract}
Embodied language grounding requires more than identifying the referent of an instruction: recovered semantics must also control the action an agent exposes. We study this missing link as a \textbf{semantic--action gap}, where instruction semantics are recoverable but weakly expressed in native continuous actions. We introduce \textbf{SAT-Bench}, a fixed-observation counterfactual benchmark that holds the visual scene and agent state fixed while changing only instruction semantics. On LIBERO target-name and pixel-grounded relation swaps, target recovery reaches 100.0\% and 95.8\%, whereas OpenVLA action sensitivity remains only 6.8\% and 7.7\%. The gap persists across 1,000 additional compositional and temporal/procedural counterfactuals, with overall action sensitivity of 6.1\%. Hidden-state, threshold-free, cross-policy, and rollout diagnostics further support this semantic--action transfer failure. We introduce \textbf{VISA}, a lightweight execution-time interface that converts recovered semantics into \textsc{allow}, \textsc{defer}, target-consistency, and verified-selection decisions. VISA reduces invalid-instruction blind execution from 92.7\% to 2.8\% while preserving 94.0\% of normal commands, and verified selection further improves target-consistent action exposure without updating the underlying policy. Overall, embodied language evaluation should measure semantic--action transfer, not semantic parsing alone.

\end{abstract}

\section{Introduction}
\label{sec:intro}

Multimodal language-grounding evaluations often ask whether a model identifies the correct object, relation, or instruction state~\citep{goyal2017making,agrawal2018dont,thrush2022winoground,su2024actplan}. Embodied language agents add an action-interface requirement: recovered semantics should affect whether and how an action is exposed. However, a vision-language-action (VLA) policy~\citep{kim2024openvla,octo2024} may recover the requested target and still move toward a plausible object from scene priors, or may parse a prohibited command and still expose an executable motion. We therefore ask an action-boundary question:
\emph{when the world is fixed, does changing instruction semantics change the native continuous action accordingly?}

\begin{figure}[t]
\centering
\includegraphics[width=\linewidth]{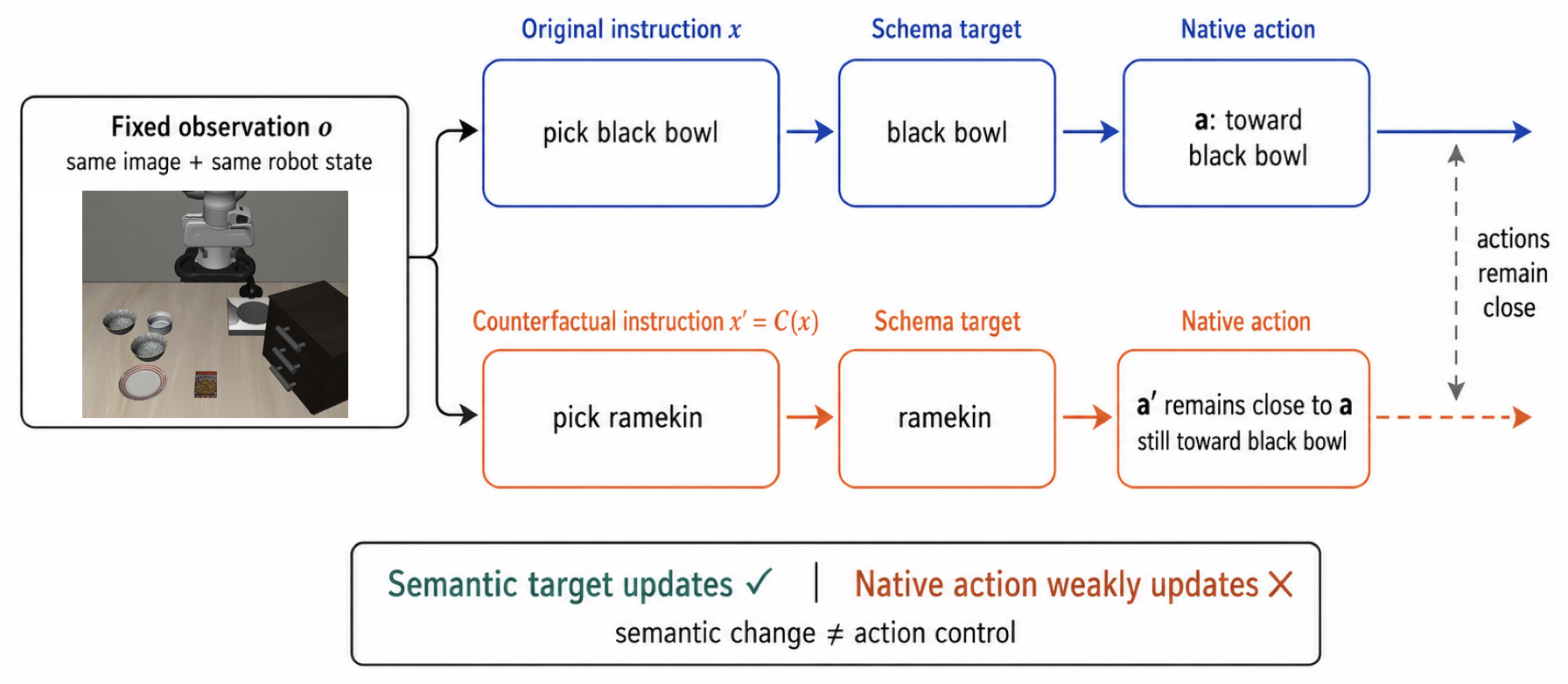}
\caption{
Fixed-observation counterfactual protocol. The observation and agent state are fixed while only the instruction semantics changes.
}
\label{fig:framework}
\end{figure}

We call this interface-level requirement \emph{semantic--action transfer} (SAT): recovered instruction semantics should affect whether and how a native action is exposed. When semantics are recoverable but remain weakly expressed in the policy's native seven-degree-of-freedom (7-DoF) action, we say the policy exhibits a \emph{semantic--action gap} (SAG). To measure this gap, we use a \emph{fixed-observation counterfactual protocol}: holding the observation and agent state fixed, we change only instruction semantics and test whether the exposed native action changes accordingly. Figure~\ref{fig:framework} illustrates this protocol, which separates semantic recoverability from semantic control of action. This distinction is the central thesis of the paper: semantic availability is not semantic action control, so embodied language evaluation should measure semantic-to-action transfer rather than semantic parsing or task success alone.

This paper makes this action-boundary problem measurable in two steps: first through a benchmark for diagnosing semantic--action transfer, and then through an execution-time interface that regulates action exposure. We introduce \textsc{SAT-Bench} (\emph{Semantic--Action Transfer Benchmark}), a fixed-observation counterfactual benchmark spanning referential, relational, compositional, temporal/procedural, invalid, prohibitive, and paraphrase-preserving instruction changes. On LIBERO~\citep{liu2023libero} target-name and pixel-grounded relation swaps, Qwen2.5-VL~\citep{bai2025qwen25vl} recovers the intended target in 100.0\% and 95.8\% of cases, yet OpenVLA action sensitivity is only 6.8\% and 7.7\%, respectively. The gap persists across 1,000 additional compositional and temporal/procedural counterfactuals, with overall action sensitivity of 6.1\%. Hidden-state probes, threshold-free analyses, schema--action attribution, and short-horizon rollouts test alternative explanations based on absent semantic decodability, threshold choice, grounding failure, and one-step artifacts. Importantly, probing is used only as evidence that semantic information is decodable from model representations, not that the action decoder causally relies on it. The same action-boundary failure also appears for non-executable language, where invalid or prohibited instructions can still trigger blind native execution.

These diagnostics motivate an explicit semantic--action interface: recovered semantics should not only be readable, but should regulate which actions are exposed. We therefore introduce \method{} (\emph{Verifiable Interface for Semantic Action}), a lightweight execution-time interface that extracts a compact instruction-conditioned schema, checks admissibility, passes through valid native actions, defers invalid ones, and flags target-inconsistent actions. A verified-selection extension further uses the recovered semantics to select among candidate actions before exposure. \method{} does not update the underlying VLA policy or claim to solve low-level or long-horizon control; its intervention target is the action boundary itself. Empirically, \method{} reduces invalid-instruction blind execution from 92.7\% to 2.8\% while preserving 94.0\% of normal commands, and verified selection further improves target-consistent action exposure. Thus, \textsc{SAT-Bench} diagnoses semantic--action transfer, while \method{} operationalizes the same distinction at execution time.

Our contributions are four-fold:
\begin{itemize}[topsep=0pt, itemsep=2pt, parsep=0pt, leftmargin=1em]
\item a formulation of \emph{semantic--action transfer} and \emph{semantic--action gaps} as interface-level requirements for embodied language grounding;
\item \textsc{SAT-Bench}, a release-ready fixed-observation counterfactual benchmark spanning referential, relational, compositional, temporal/procedural, invalid, prohibitive, and paraphrase-preserving semantics;
\item diagnostic evidence that instruction semantics can be recoverable externally and linearly decodable from OpenVLA representations while remaining weakly expressed in native actions, together with behavioral tests of alternative explanations;
\item \method{}, a schema-based execution-time semantic--action interface for allow, defer, target-consistency, and verified-selection decisions without updating or replacing the underlying low-level policy.
\end{itemize}

\section{Related Work}
\label{sec:related}

\paragraph{Grounded language evaluation, counterfactuals, and probing.}
Multimodal grounding work tests whether models recover intended objects, relations, or scene states from language and perception. Counterfactual and contrastive evaluations further ask whether predictions rely on intended semantic variables rather than dataset priors or superficial correlations~\citep{goyal2017making,agrawal2018dont,thrush2022winoground}. Related embodied and procedural benchmarks study planning, grounding, scene consistency, and robustness under semantic perturbations~\citep{su2024actplan,chakraborty2025heal}. We target a narrower action-interface question: holding the observation and agent state fixed, does changing only the instruction induce the corresponding change in the exposed action? We also draw on representation probing, which tests whether task-relevant information is present in intermediate states~\citep{alain2016understanding,hewitt2019structural,belinkov2022survey}; in our setting, probing helps separate absent semantic recovery from failed transfer into action~\citep{zhang2025knowing}.

\paragraph{Embodied language agents and VLA benchmarks.}
Recent embodied foundation models connect visual observations, language instructions, and actions through imitation learning, multimodal pretraining, and large robot datasets~\citep{brohan2022rt1,driess2023palme,zitkovich2023rt2,oneill2023openx}. OpenVLA and Octo provide open-source VLA policies that enable controlled inspection of language-conditioned action behavior~\citep{kim2024openvla,octo2024}, while LIBERO offers language-conditioned manipulation tasks for transfer evaluation~\citep{liu2023libero}. These works primarily evaluate task success, generalization, or imitation quality. Our focus is complementary: we use native actions as a readout of grounded-language transfer rather than as the sole target of control optimization.

\paragraph{Invalid instructions, structured semantics, and action interfaces.}
A related line of work studies whether embodied agents can reject false-premise commands, infeasible requests, unsafe instructions, or scene-task inconsistencies~\citep{hsieh2025dowhat,chakraborty2025heal,son2025safel}. High-level language-to-robot systems combine language reasoning with affordances, feedback, or programmatic primitives to improve executable planning~\citep{ahn2022saycan,huang2022inner,liang2022code}. Robotic language understanding has long used structured semantic representations to connect commands with objects, relations, actions, and constraints~\citep{tellex2011understanding,thomason2015learning,thomason2019improving}; modern VLMs such as Qwen2.5-VL provide image-conditioned parsing and grounding~\citep{bai2025qwen25vl}. VISA builds on this interface view, but uses schemas for evaluation and exposure decisions: whether recovered semantics should allow, defer, or flag an action.

\section{Semantic--Action Transfer Framework}
\label{sec:framework}

We formalize embodied instruction following as an action-interface problem with two separable requirements: the relevant semantic state must be recoverable, and the exposed continuous action must appropriately reflect that state. The goal is not to optimize robot motion, but to evaluate whether recovered semantics reach the native action interface. We first define semantic--action transfer under fixed observations, then introduce \textsc{SAT-Bench} interventions, and finally define measurements for action sensitivity, non-execution, and target consistency.

\subsection{Interface Requirement under Fixed Observations}
\label{sec:problem_formulation}

Let $o$ denote the visual observation together with the robot state, $x$ a natural-language instruction, and $a=f(o,x)$ the native continuous action predicted by a VLA policy. We study an interface-level requirement: if instruction semantics change while the world state is fixed, the exposed action should respond in the corresponding semantic direction, either by redirecting motion or by inhibiting action exposure when execution is inadmissible.

A counterfactual instruction $x'=C(x)$ changes the requested semantics while keeping $o$ fixed, yielding $a'=f(o,x')$. The fixed-observation design is a measurement protocol rather than the definition of the failure itself. If both the observation and instruction change, an action difference could be caused by language, visual features, object layout, or scene priors. By holding $o$ fixed, the protocol isolates the instruction-conditioned component of action selection.

This formulation separates two questions often conflated in embodied evaluation: whether the relevant semantics are recoverable from $(o,x)$, and whether the native action boundary reflects those semantics. We call the desired interface property \emph{semantic--action transfer} (SAT). When semantic recovery is high but corresponding action control is weak, the policy exhibits a \emph{semantic--action gap} (SAG). Thus, SAG is not a claim that the model lacks language understanding, nor that a particular internal representation is causally ignored; it is a behavioral claim that recovered semantics are weakly expressed at the native action boundary.

\subsection{\textsc{SAT-Bench} Interventions}
\label{sec:setup_splits}

\textsc{SAT-Bench} instantiates the fixed-observation protocol through controlled instruction interventions while the observation and robot state remain fixed. We define the intervention family
\begin{equation}
\mathcal{C} =
\{
C_{\mathrm{name}},
C_{\mathrm{rel}},
C_{\mathrm{comp}},
C_{\mathrm{proc}},
C_{\mathrm{invalid}},
C_{\mathrm{neg}},
C_{\mathrm{para}}
\},
\end{equation}
with $C \in \mathcal{C}$ probing a distinct semantic-to-action property.

$C_{\mathrm{name}}$ replaces the explicitly named target while the scene is fixed, e.g., ``pick up the black bowl'' becomes ``pick up the ramekin,'' testing whether target identity redirects the exposed action.
$C_{\mathrm{rel}}$ removes the target name and specifies the target through a spatial relation to an anchor object, testing whether grounded relational semantics reach the same action boundary.
$C_{\mathrm{comp}}$ changes compositional target semantics through attribute--relation combinations or multiple simultaneous constraints, testing whether richer conjunctions alter action exposure.
$C_{\mathrm{proc}}$ changes temporal or procedural semantics, including the required first subgoal, action ordering, or procedural constraints, testing whether the exposed action reflects the currently required step.
$C_{\mathrm{invalid}}$ converts executable commands into blank, impossible, ambiguous, or unsupported instructions, testing whether non-executable semantics inhibit action exposure.
$C_{\mathrm{neg}}$ constructs prohibited commands such as \texttt{do\_not}, \texttt{avoid}, \texttt{leave\_alone}, and \texttt{except}, testing whether language can block an otherwise plausible action.
Finally, $C_{\mathrm{para}}$ produces semantically equivalent paraphrases and serves as a control for wording-induced action variation.

Together, these interventions test three interface behaviors required for semantic--action transfer: redirecting actions when executable semantics change, inhibiting actions when execution becomes inadmissible, and calibrating the amount of native action variation induced by wording alone.

\subsection{Measurement Stack}
\label{sec:setup_action_sensitivity}

We use a three-part measurement stack. \textbf{Action-sensitivity metrics} test whether an executable semantic intervention changes the native action beyond wording-induced variation. \textbf{Non-execution metrics} test whether non-executable semantics suppress action exposure. \textbf{Target-consistency metrics} test whether an exposed action is directed toward the recovered target rather than a competing target. Together, these measurements separate three interface questions: whether actions respond to semantic changes, whether inadmissible actions are withheld, and whether exposed actions are semantically directed.

\paragraph{Action-sensitivity metrics.}
For executable semantic interventions, we measure native action sensitivity using a normalized action difference:
\begin{equation}
    D(a,a') =
    \left\|
    (a-a') \oslash \sigma_a
    \right\|_2,
\end{equation}
where $\sigma_a$ is the per-dimension action standard deviation for the evaluated policy. An action is counted as sensitive if
\begin{equation}
    D(a,a') > \tau_s,
\end{equation}
where $\tau_s$ is calibrated separately for each policy and evaluation setting as the 95th percentile of action deltas from semantics-preserving $C_{\mathrm{para}}$ controls. These controls keep the observation and task semantics fixed while varying only wording. Target-changing counterfactuals are never used to set $\tau_s$; calibration and evaluation therefore share the observation distribution but are disjoint in perturbation type. We report the exact setting-specific thresholds in the appendix.

Given a task-appropriate semantic recovery signal $\mathrm{SemSens}$, we quantify the semantic--action gap as
\begin{equation}
    \mathrm{SAG}
    =
    \mathrm{SemSens}
    -
    \mathrm{ActionSens}.
\end{equation}
For target-name swaps, $\mathrm{SemSens}$ is schema sensitivity; for oracle relation swaps, $\mathrm{SemSens}=1$; and for pixel-grounded relation swaps, $\mathrm{SemSens}$ is grounding accuracy over all examples or $1$ on the grounding-correct subset. For richer compositional and procedural interventions, $\mathrm{SemSens}$ is defined by recovery of the intervention-specific target or required subgoal. To avoid making the diagnosis depend on a single calibrated cutoff, we additionally report target-versus-control AUC over the full action-delta distribution.

\paragraph{Non-execution metrics.}
For inadmissible instructions, the desired interface behavior is non-execution rather than a different target-directed motion. Let $a_{\mathrm{orig}}=f(o,x)$ be the action under the original valid instruction and $a_{\mathrm{inv}}=f(o,x_{\mathrm{invalid}})$ the action under an invalid perturbation. Using the matched policy- and setting-specific threshold $\tau_s$, we define
\begin{equation}
\begin{aligned}
\mathrm{BlindExec}
&=
\mathbb{I}
\!\left[
D(a_{\mathrm{orig}},a_{\mathrm{inv}})<\tau_s
\right],\\
\mathrm{ActionInhib}
&=
1-\mathrm{BlindExec}.
\end{aligned}
\end{equation}
Blind execution measures whether an invalid instruction still exposes a native action close to the original executable command. Action inhibition measures whether non-executable semantics suppress that exposure. At the schema level, an invalid instruction is rejected if the execution decision is a safe deferral such as \texttt{ask}, \texttt{hold}, \texttt{abort}, or \texttt{target-not-found}. We report invalid safe deferral together with normal-command pass rate so that a trivial all-stop system cannot score well by refusing every instruction.

\paragraph{Target-consistency metrics.}
\label{sec:setup_actcheck}

Action sensitivity only asks whether an action changes; it does not determine whether the changed action is directed toward the intended target. We therefore define \actcheck{}, a target-consistency diagnostic. Let $p_e$ be the end-effector position, $p_t$ the intended target position, $p_w$ a competing wrong or original target, and $v=a_{1:3}$ the translational action. We compute
\begin{equation}
c_t=\cos(v,p_t-p_e),
\qquad
c_w=\cos(v,p_w-p_e).
\end{equation}
With margin $\delta_{\mathrm{ac}}$, the action is target-aligned if $c_t-c_w>\delta_{\mathrm{ac}}$, wrong-target if $c_w-c_t>\delta_{\mathrm{ac}}$, and ambiguous otherwise.

Crossing this label with schema correctness yields \emph{schema--action attribution}, which separates semantic-recovery failures from semantic-to-action transfer failures. A wrong schema with a wrong action is primarily a recoverability failure; a correct schema with a wrong or ambiguous action is an interface-transfer failure; and a correct schema with an aligned action is the desired case. \actcheck{} is therefore an action-boundary diagnostic rather than a measure of full manipulation success.

\subsection{From Diagnostics to \method{}}
\label{sec:why_interface}

These measurements localize failures at the action boundary. If semantics are unrecoverable, the bottleneck lies in perception or language grounding; if semantics are recoverable but executable semantic changes are weakly reflected in native actions, the bottleneck is semantic-to-action transfer; if inadmissible instructions still expose actions, the bottleneck is admissibility; and if exposed actions point toward the wrong target, the bottleneck is target consistency. These failure modes motivate \method{}: an execution-time interface that converts recovered semantics into explicit action-exposure decisions---allowing admissible native actions, deferring inadmissible instructions, flagging target-inconsistent actions, or selecting among verified action proposals---without updating or replacing the underlying low-level policy.
\section{Diagnostic Evidence for the Semantic--Action Gap}
\label{sec:diagnostic_results}

We ask three questions: whether recoverable semantics control native actions,
whether simpler explanations account for the observed gap, and whether the
same boundary extends beyond simple target changes. Unless otherwise specified,
the action policy is OpenVLA and the schema or grounding probe is
Qwen2.5-VL-7B-Instruct. We additionally evaluate OpenVLA-LIBERO90 and
Octo-small-1.5, and use BridgeData V2 as a real-robot instruction-change
diagnostic rather than a strict object-grounded target-swap benchmark.
External probes measure semantic recoverability only; hidden-state probes
measure linear decodability rather than causal use by the action decoder.
The full diagnostic pipeline is provided in
Appendix~\ref{app:diagnostic_pipeline}.


\begin{table*}[t]
\centering
\scriptsize
\setlength{\tabcolsep}{5.2pt}
\renewcommand{\arraystretch}{1.04}

\begin{tabular*}{0.96\textwidth}{
@{\extracolsep{\fill}}
lllrrrrr
@{}}
\toprule
\textbf{Policy}
& \textbf{Domain}
& \textbf{Intervention}
& $\boldsymbol{N}$
& \textbf{Sem.}
& \textbf{Act.}
& \textbf{Gap}
& \textbf{ActCheck} \\
\midrule

OpenVLA base
& LIBERO
& target-name
& 600 & 100.0 & 6.8 & 93.2 & -- \\

OpenVLA-LIBERO90
& LIBERO
& target-name
& 600 & 100.0 & 5.7 & 94.3 & -- \\

Octo-small-1.5
& LIBERO
& target-name
& 600 & 100.0 & 42.7 & 57.3 & -- \\

\addlinespace[1.5pt]

OpenVLA base
& LIBERO
& pixel relation
& 600 & 95.8 & 7.7 & 88.1 & 45.5 \\

Octo-small-1.5
& LIBERO
& pixel relation
& 600 & 95.8 & 37.5 & 58.3 & 42.8 \\

\addlinespace[1.5pt]

Octo-small-1.5
& BridgeData V2
& target/task
& 165 & 100.0 & 66.1 & 33.9 & limited \\

\bottomrule
\end{tabular*}

\caption{
Semantic-to-action transfer across policies and domains.
Sem. denotes semantic recovery, Act. denotes native action sensitivity,
and Gap is Sem. minus Act.; numeric values except $N$ are percentages.
}
\label{tab:cross_policy_diagnostics}
\end{table*}


\subsection{Recoverable Semantics Weakly Control Native Actions}
\label{sec:diag_main_gap}

On target-name swaps, the schema target changes in 100.0\% of cases while
OpenVLA action sensitivity is only 6.8\%. On pixel-grounded relation swaps,
Qwen2.5-VL recovers the intended target from the image and relation in 95.8\%
of cases, yet OpenVLA action sensitivity remains 7.7\%. The latter setting is
especially diagnostic because the target name is absent from the instruction.
Thus, high semantic recovery can coexist with weak expression of the same
change at the native action boundary.

Table~\ref{tab:cross_policy_diagnostics} shows that the phenomenon is graded
rather than specific to one OpenVLA checkpoint. OpenVLA-LIBERO90 exhibits a
similar gap, while Octo-small-1.5 is substantially more action-sensitive.
BridgeData V2 shows still greater responsiveness, but should be interpreted
only as a real-robot instruction-change diagnostic because it lacks the object
metadata required for strict target swaps.

\begin{figure}[t]
\centering
\includegraphics[width=0.92\linewidth]{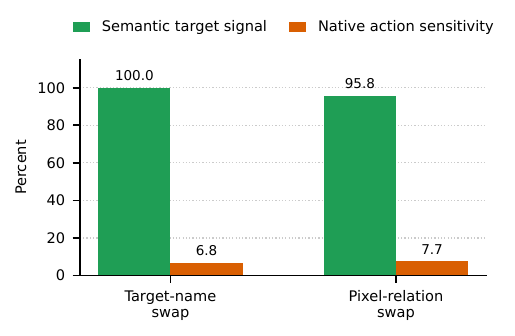}
\caption{
Main semantic--action gap on OpenVLA counterfactuals:
semantic target signals remain high while native action sensitivity is low.
}
\label{fig:main_gap_chart}
\end{figure}


\subsection{Ruling Out Alternative Explanations}
\label{sec:diag_threshold_quadrant}

A low action-sensitivity score admits several alternative explanations.
Table~\ref{tab:diagnostic_alternatives} summarizes the diagnostic used to test
each one.

\begin{table}[H]
\centering
\scriptsize
\setlength{\tabcolsep}{4pt}
\renewcommand{\arraystretch}{1.04}

\begin{tabularx}{\linewidth}{@{}
>{\raggedright\arraybackslash}X
>{\raggedright\arraybackslash}p{0.39\linewidth}
@{}}
\toprule
\textbf{Alternative explanation}
& \textbf{Diagnostic} \\
\midrule

Semantics absent from VLA representations
& Hidden-state probe \\

Threshold artifact
& Threshold-free AUC \\

Grounding-only failure
& Schema--action attribution \\

Overlapping target directions
& Spatial subsets \\

First-step irrelevance
& Short-horizon rollouts \\

Simple-semantics only
& Richer semantic splits \\

\bottomrule
\end{tabularx}

\caption{
Alternative explanations and the diagnostics used to test them.
}
\label{tab:diagnostic_alternatives}
\end{table}

\paragraph{Internal decodability.}
\label{sec:hidden_probe_main}

A natural alternative is that only the external Qwen probe recovers the target
while OpenVLA itself lacks the corresponding information. Linear probes show
otherwise: in the full pixel-relation condition, target schema sensitivity is
1.000 while decoded action sensitivity is 0.094; on an image-dependent subset
where language-only or shuffled-image controls fail, the corresponding values
are 1.000 and 0.060. Full layer, masking, shuffled-image, and random-label
controls are reported in the appendix.

These results establish linear decodability, not causal reliance by the action
decoder. Preliminary activation patching likewise has limited effect
(Appendix~\ref{app:activation_patching}), so we do not claim to identify the
complete internal causal mechanism.

\paragraph{Metric and grounding robustness.}
Threshold-free target-versus-paraphrase comparisons yield AUCs of 57.3 and
58.3 for OpenVLA target-name and pixel-relation swaps, respectively. In
contrast, BridgeData V2 with Octo reaches 92.9
(Appendix~\ref{app:threshold_free}), showing that the metric can register
stronger instruction-conditioned action separation when present. Spatially
separated target subsets give the same qualitative OpenVLA result
(Appendix~\ref{app:angle_separated}).

Schema--action attribution further separates grounding from transfer failures.
On the 600 pixel-grounded relation examples, 44.0\% have the correct recovered
target but a wrong or ambiguous native action, whereas only 4.2\% fall in
schema-wrong regions.

\begin{table}[t]
\centering
\scriptsize
\setlength{\tabcolsep}{5pt}
\renewcommand{\arraystretch}{1.04}

\begin{tabular*}{\linewidth}{
@{\extracolsep{\fill}}
lrr
@{}}
\toprule
\textbf{Schema/action case}
& \textbf{Count}
& \textbf{Rate (\%)} \\
\midrule

Correct target + aligned action
& 311 & 51.8 \\

\textbf{Correct target + wrong/ambig. action}
& \textbf{264} & \textbf{44.0} \\

Wrong target
& 25 & 4.2 \\

\bottomrule
\end{tabular*}

\caption{
Schema--action attribution on 600 pixel-grounded relation examples.
}
\label{tab:schema_action_quadrant}
\end{table}

\paragraph{Short-horizon behavioral relevance.}
\label{sec:robustness_diagnostics}

One-step sensitivity is not itself a task-success metric. Across 300 paired
OpenVLA-LIBERO90 rollouts of $K=20$ steps, raw counterfactual instructions do
not reliably redirect trajectories. However, first-step \actcheck{} labels
significantly stratify later behavior: target-aligned first actions yield
higher counterfactual-target approach and target preference than ambiguous or
wrong-target actions. Full rollout statistics are reported in
Appendix~\ref{app:actcheck_predictive_rollout_300}. We therefore use
\actcheck{} as a short-horizon predictive diagnostic rather than as evidence
of full task success.


\subsection{Beyond Simple Target Changes}
\label{sec:diag_extensions}
\label{sec:diag_richer_semantics}
\label{sec:diag_negation}

We next test semantic breadth along two complementary axes. First, we expand
\textsc{SAT-Bench} with 1,000 fixed-observation counterfactuals covering
compositional and temporal/procedural semantics.

\begin{table}[t]
\centering
\scriptsize
\setlength{\tabcolsep}{5pt}
\renewcommand{\arraystretch}{1.05}

\begin{tabular*}{\linewidth}{
@{\extracolsep{\fill}}
lrrr
@{}}
\toprule
\textbf{Split}
& $\boldsymbol{N}$
& \textbf{Act. sens. (\%)}
& \textbf{SAG (\%)} \\
\midrule

Compositional
& 600 & 7.0 & 93.0 \\

Temporal/procedural
& 400 & 4.8 & 95.3 \\

\midrule

\textbf{Overall}
& \textbf{1,000}
& \textbf{6.1}
& \textbf{93.9} \\

\bottomrule
\end{tabular*}

\caption{
Semantic--action transfer on richer \textsc{SAT-Bench} interventions.
}
\label{tab:richer_semantics}
\end{table}

The gap remains large: action sensitivity is 7.0\% for compositional changes
and 4.8\% for temporal/procedural changes. The phenomenon therefore extends
beyond explicit object replacement and simple spatial relations to conjunctions
of semantic constraints and changes in the required first subgoal or action
ordering.

Second, the same boundary appears when semantics should inhibit rather than
redirect action. OpenVLA blindly executes 92.7\% of templated invalid
instructions. On 800 Octo negation/prohibition variants, only 29.8\% are
inhibited, leaving 70.2\% blind execution.

\begin{figure}[t]
\centering
\includegraphics[width=0.94\linewidth]{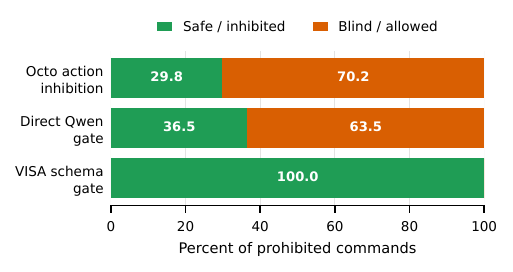}
\caption{
Non-executable language stress test on prohibited commands.
}
\label{fig:negation_stress_main}
\end{figure}

Because benchmark-label errors could themselves create an apparent
semantic--action gap, we additionally perform blind external validation with
three non-author annotators on 300 randomized items, yielding 900 annotation
slots. Table~\ref{tab:external_validation_main} summarizes agreement with the
benchmark labels.

\begin{table}[t]
\centering
\scriptsize
\setlength{\tabcolsep}{4.2pt}
\renewcommand{\arraystretch}{1.05}

\begin{tabular*}{\linewidth}{
@{\extracolsep{\fill}}
lccc
@{}}
\toprule
\textbf{Field}
& \textbf{Agreement}
& \textbf{Fleiss' $\boldsymbol{\kappa}$}
& \textbf{Maj. acc.} \\
\midrule

Executability
& 0.964 & 0.825 & 1.000 \\

Intended target
& 0.987 & 0.975 & 0.997 \\

First required subgoal
& \textbf{0.691}
& \textbf{0.524}
& \textbf{0.770} \\

Robot decision
& 0.978 & 0.916 & 1.000 \\

\bottomrule
\end{tabular*}

\caption{
Blind external validation with three non-author annotators.
The lower first-subgoal agreement reflects ambiguity in procedural decomposition.
}
\label{tab:external_validation_main}
\end{table}

Agreement is high for executability, intended target, and robot decision.
First-subgoal agreement is lower, reflecting the greater ambiguity of
procedural decomposition; we therefore treat temporal/procedural results as
controlled diagnostics rather than exhaustive procedural-language evaluation.
Full agreement statistics and audit details are reported in
Appendix~\ref{app:hard_invalid_audit}.

Together, these results localize a broader action-boundary failure: semantics
can be recoverable through external grounding or linearly decodable from VLA
representations, yet remain weakly reflected when action should be redirected,
inhibited, or made target-consistent. This diagnosis motivates the
execution-time interface introduced next.

\FloatBarrier
\section{\method{}: A Semantic--Action Interface for Action Exposure}
\label{sec:visa}

\begin{figure}[t]
\centering
\includegraphics[width=0.92\linewidth,trim=0 2 0 2,clip]
{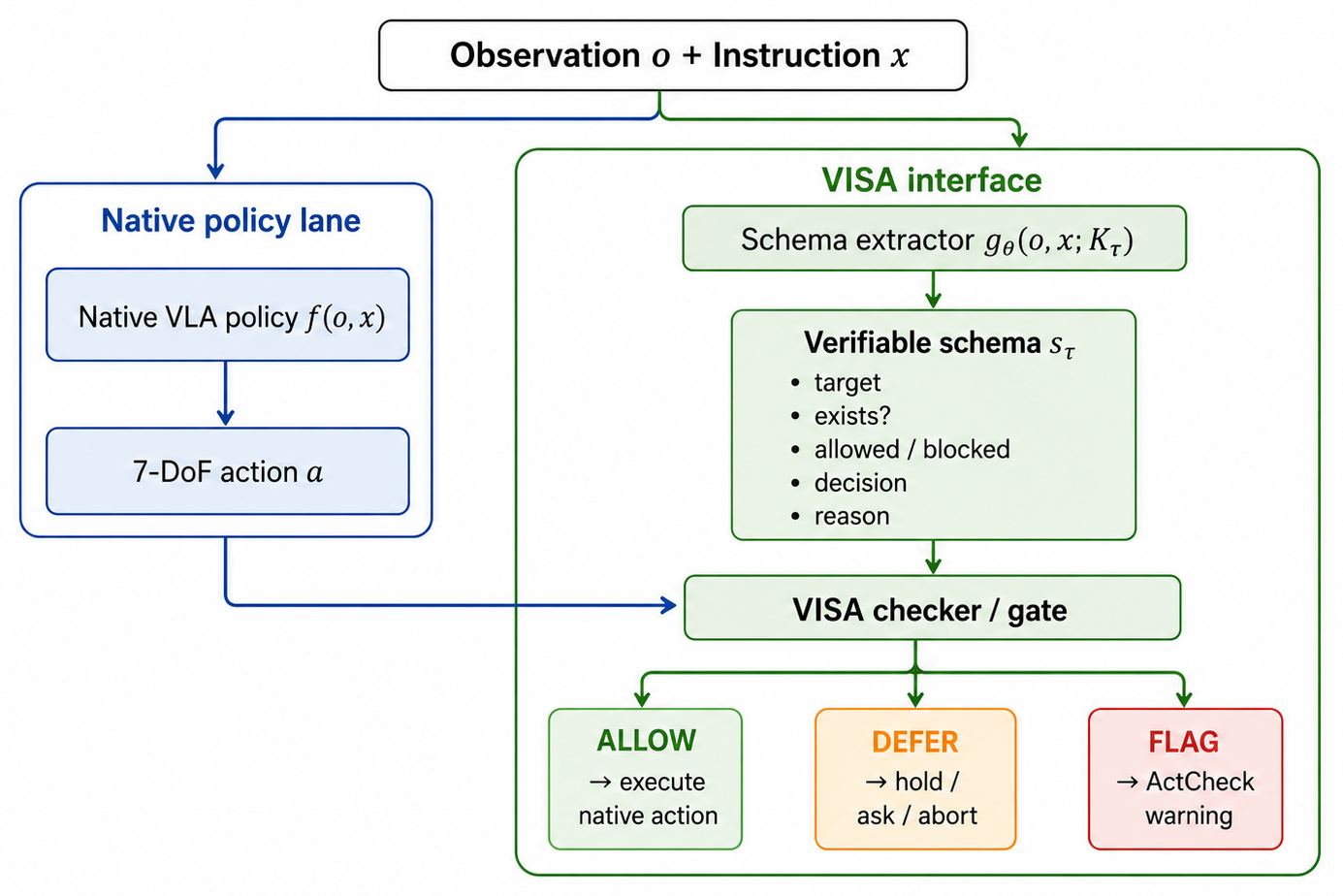}
\caption{
\method{} semantic--action interface: schema extraction,
admissibility checking, and action exposure.
}
\label{fig:visa_interface}
\end{figure}

Section~\ref{sec:diagnostic_results} localizes semantic--action failures at the
action boundary: recovered semantics may fail to redirect native actions,
inadmissible instructions may still expose executable motions, and
target-correct schemas may still yield wrong-target or ambiguous actions.
\method{} responds with an explicit execution-time interface that makes
recovered semantics inspectable and uses them to regulate action exposure.
It passes through admissible native actions, defers inadmissible instructions,
and flags target-inconsistent actions before execution.
Figure~\ref{fig:visa_interface} summarizes the interface.

\subsection{Schema, Admissibility, and Action Exposure}
\label{sec:visa_components}

\paragraph{Schema extraction.}
For a task family $\tau$, \method{} extracts an instruction-conditioned schema
\begin{equation}
    s_\tau=(t,e,A^+,A^-,u,d,r),
\end{equation}
where $t$ denotes the target, $e$ target existence, $A^+$ and $A^-$ allowed
and blocked symbolic actions, $u$ the next symbolic action, $d$ the execution
decision, and $r$ a reason. Task families expose only the fields required by
their semantics: relation tasks additionally represent anchor and relation
variables, while temporal/procedural tasks represent the required subgoal or
ordering constraint.

Schemas are specified once per task family rather than annotated per instance.
Table~\ref{tab:visa_schema_effort} summarizes the corresponding design effort.
Across the families studied here, schema construction requires low-to-medium
manual effort and no instance-level schema labels.

\begin{table}[!ht]
\centering
\scriptsize
\setlength{\tabcolsep}{4pt}
\renewcommand{\arraystretch}{1.05}

\begin{tabularx}{\linewidth}{@{}
>{\raggedright\arraybackslash}p{0.26\linewidth}
>{\raggedright\arraybackslash}X
cc
@{}}
\toprule
\textbf{Task family}
& \textbf{Additional semantic fields}
& \textbf{Effort}
& \textbf{Inst. labels} \\
\midrule

Target-name
& target / existence
& low
& no \\

Relation
& anchor / relation
& low--med.
& no \\

Invalid / prohibition
& blocked actions / decision
& low
& no \\

Compositional
& attributes / constraints
& low--med.
& no \\

Temporal / procedural
& subgoal / ordering
& medium
& no \\

\bottomrule
\end{tabularx}

\caption{
Task-family schema design requirements.
Schemas are specified once per family rather than annotated per instance.
}
\label{tab:visa_schema_effort}
\end{table}

\paragraph{Do the schema fields matter?}
We evaluate this question with a controlled 300-example target-field ablation.
Table~\ref{tab:visa_target_ablation_main} reports the change in candidate recall
between the complete schema and a variant with the target field removed.

\begin{table}[!ht]
\centering
\scriptsize
\setlength{\tabcolsep}{5pt}
\renewcommand{\arraystretch}{1.05}

\begin{tabular*}{\linewidth}{
@{\extracolsep{\fill}}
rrrr
@{}}
\toprule
\textbf{Budget}
& \textbf{$\Delta$ Recall}
& \textbf{95\% CI}
& \textbf{McNemar $p$} \\
\midrule

2
& \textbf{+0.087}
& [0.040, 0.137]
& 0.0005 \\

4
& \textbf{+0.080}
& [0.030, 0.130]
& 0.0027 \\

\bottomrule
\end{tabular*}

\caption{
Target-field ablation. Positive $\Delta$ Recall denotes the recall advantage
of the complete schema over removing the target field.
}
\label{tab:visa_target_ablation_main}
\end{table}

Removing the target field significantly reduces candidate recall at both
proposal budgets, showing that the explicit semantic state contributes to
candidate verification rather than serving only as a descriptive explanation.
At the same time, a strong unstructured proposal-selection baseline remains
competitive, so we do not claim that structured schemas are uniquely
responsible for final action quality. Their demonstrated role is to expose
auditable semantic variables that can be checked explicitly before action
exposure.

\paragraph{Admissibility and target consistency.}
The admissibility checker enforces instruction-conditioned constraints over
$s_\tau$: blank or impossible instructions should not trigger target-directed
motion, and prohibited instructions should not execute blocked actions. The
base interface is
\begin{equation}
\pi_{\mathrm{VISA}}(o,x)=
\begin{cases}
(f(o,x),\textsc{ALLOW}), & \mathrm{admissible}(s_\tau),\\
(\varnothing,\textsc{DEFER}(r)), & \mathrm{otherwise}.
\end{cases}
\end{equation}
Thus, admissible native actions pass through unchanged, whereas inadmissible
instructions defer execution. For allowed actions, \actcheck{} applies the
target-consistency test from Section~\ref{sec:setup_actcheck}; wrong-target or
ambiguous actions are surfaced as \textsc{FLAG} decisions before exposure.

\paragraph{Verified proposal selection.}
A \textsc{FLAG} identifies an inconsistent native action but does not itself
provide an alternative. We therefore consider \method{}-Rerank, an optional
proposal-and-verification extension: candidate actions are checked against the
recovered semantic constraints, the highest-ranked verified candidate is
exposed, and the interface defers if no candidate passes verification. This
extends \method{} from detecting target-inconsistent exposure to selecting
among candidate actions; Section~\ref{sec:visa_eval} evaluates both
verifier-based exposure quality and independent behavioral outcomes.

\paragraph{Scope.}
\method{} regulates action exposure rather than repairing policy weights. It
does not update the underlying VLA, synthesize grasps, plan complete
trajectories, or certify long-horizon task success. \textsc{SAT-Bench}
therefore diagnoses semantic--action transfer, while \method{} operationalizes
the same distinction at execution time by deciding whether an action should be
allowed, deferred, flagged, or selected.
\FloatBarrier
\section{Evaluating \method{} as an Action-Exposure Interface}
\label{sec:visa_eval}

We evaluate whether \method{} can (i) defer inadmissible instructions without
collapsing normal-command utility and (ii) improve target-consistent action
exposure when native actions are unreliable. These are action-boundary outcomes,
not claims of policy repair or full task success.


\subsection{Deferring Invalid Commands while Preserving Utility}
\label{sec:visa_safe_deferral}

The non-execution diagnostics in Section~\ref{sec:diag_negation} show that
native policies can expose plausible motions even when execution should be
deferred. On templated invalid instructions, \method{} raises safe deferral to
97.2\%, reducing blind execution from 92.7\% to 2.8\%. Under 800
negation/prohibition variants, Octo inhibits only 29.8\% of prohibited commands
and a direct binary Qwen gate safely defers 36.5\%, whereas the structured
\method{} gate reaches 100.0\%.

\begin{figure}[t]
\centering
\includegraphics[width=0.90\linewidth]
{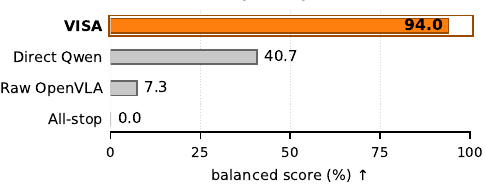}
\caption{
Deferral--utility balance: the minimum of invalid safe deferral and normal
pass-through.
}
\label{fig:visa_safety_utility}
\end{figure}

Safe deferral alone is insufficient because an all-stop system can appear safe
by rejecting every instruction. Figure~\ref{fig:visa_safety_utility} therefore
reports the minimum of invalid safe deferral and normal-command pass-through.
\method{} scores 94.0, compared with 40.7 for Direct Qwen, 7.3 for raw OpenVLA,
and 0.0 for all-stop.

\paragraph{Generalization beyond templates.}
To test whether these gains rely on surface-form regularities, we evaluate 300
non-template invalid instructions spanning paraphrased prohibitions, implicit
impossibilities, ambiguous references, instruction conflicts, and distractor
non-commands.

\begin{table}[t]
\centering
\scriptsize
\setlength{\tabcolsep}{4.2pt}
\renewcommand{\arraystretch}{1.05}

\begin{tabular*}{\linewidth}{
@{\extracolsep{\fill}}
lrrr
@{}}
\toprule
\textbf{Method}
& \textbf{Safe def. $\uparrow$}
& \textbf{Normal pass $\uparrow$}
& \textbf{Invalid allow $\downarrow$} \\
\midrule

Template heuristic
& 0.123 & -- & 0.877 \\

Direct Qwen
& 0.870 & 0.982 & 0.130 \\

\method{} gate
& 0.780 & 0.940 & 0.220 \\

Hybrid
& 1.000 & 0.922 & 0.000 \\

\bottomrule
\end{tabular*}

\caption{
Generalization to 300 non-template invalid instructions.
}
\label{tab:visa_hard_invalid_main}
\end{table}

The template heuristic collapses on this split, confirming that the result is
not explained by reusing surface patterns from the templated diagnostic.
\method{} still preserves the intended safety--utility tradeoff: it safely
defers 78.0\% of invalid commands while passing 94.0\% of normal commands.
Direct Qwen attains higher invalid-set safe deferral, whereas the hybrid
eliminates invalid allows at a modest additional utility cost. We therefore
interpret these methods as different operating points rather than claiming a
universally dominant gate.

The remaining \method{} errors are also structured. Spatial-relation normal
commands pass at 100.0\%, whereas drawer/cabinet commands pass at 72.7\%
(Appendix~\ref{app:normal_utility}). This concentration localizes the utility
loss to particular task families, suggesting that task-specific schema or
admissibility refinement is preferable to globally relaxing the gate.


\subsection{Verified Action Exposure and Behavioral Relevance}
\label{sec:visa_target_flags}
\label{sec:visa_repair_signal}

Action sensitivity does not guarantee target consistency: on pixel-grounded
relations, 54.5\% of native OpenVLA actions are target-aligned and 45.5\% are
wrong-target or ambiguous. \actcheck{} exposes the latter as pre-execution
\textsc{FLAG} decisions.

\paragraph{Verified exposure across semantic interventions.}
We next ask whether recovered semantics can do more than identify these
failures. On 600 target-swap examples, \method{}-Rerank generates candidate
actions, verifies them against the recovered target, and either exposes a
verified candidate or defers. We then apply the same proposal-and-verification
interface to the compositional and temporal/procedural splits from
Section~\ref{sec:diag_richer_semantics}.

\begin{table}[t]
\centering
\scriptsize
\setlength{\tabcolsep}{4.2pt}
\renewcommand{\arraystretch}{1.05}

\begin{tabular*}{\linewidth}{
@{\extracolsep{\fill}}
lrrr
@{}}
\toprule
\textbf{Split}
& \textbf{Native aligned}
& \textbf{Allow}
& \textbf{Aligned / allow} \\
\midrule

Target swap
& 0.682 & \textbf{0.938} & \textbf{1.000} \\

Compositional
& 0.513 & 0.890 & \textbf{1.000} \\

Temporal/procedural
& 0.590 & 0.907 & \textbf{1.000} \\

\bottomrule
\end{tabular*}

\caption{
Verified action exposure across semantic interventions.
Allow denotes non-deferred coverage; Aligned / allow denotes
target-consistent exposure among allowed actions.
}
\label{tab:visa_verified_exposure_main}

\label{tab:visa_rerank_main}
\label{tab:visa_rerank_richer_main}
\end{table}

Verified selection retains 89.0--93.8\% coverage with no verifier-detected
inconsistent exposure among allowed actions across all three settings. This
extends the action-exposure result beyond explicit target replacement: the same
recovered semantic variables that diagnose compositional and procedural
transfer failures can also constrain which candidate action is exposed.

However, these results remain verifier-based. Because \actcheck{} participates
in candidate verification, perfect alignment among exposed actions cannot by
itself establish improved downstream behavior. This motivates an evaluation
with trajectory metrics that are independent of the selection verifier.

\paragraph{Independent rollout evaluation.}
We evaluate target-swap selections using trajectory metrics that are not used
by the verifier.

\begin{table}[t]
\centering
\scriptsize
\setlength{\tabcolsep}{4.0pt}
\renewcommand{\arraystretch}{1.05}

\begin{tabular*}{\linewidth}{
@{\extracolsep{\fill}}
lrrr
@{}}
\toprule
\textbf{Method}
& \textbf{Final pref. $\uparrow$}
& \textbf{Integrated pref. $\uparrow$}
& \textbf{Wrong-object $\downarrow$} \\
\midrule

Native
& 0.0002 & 0.0178 & 0.720 \\

\method{}-Rerank
& \textbf{0.0083}
& \textbf{0.0216}
& \textbf{0.600} \\

\bottomrule
\end{tabular*}

\caption{
Independent $K=100$ rollout evaluation using metrics outside the verifier.
}
\label{tab:visa_rerank_rollout_main}
\end{table}

At $K=100$, \method{}-Rerank increases final and integrated
counterfactual-target preference while reducing wrong-object approach. A
separate $K=50$, $N=100$ evaluation shows the same qualitative trend
(Appendix~\ref{app:visa_rerank_rollouts}). Thus, the exposure improvement is
also reflected in trajectory-level metrics that are independent of the
selection verifier.

\paragraph{Environment-step pilot.}
Finally, we test selective action-boundary intervention under true environment
transitions in a 10-trial manipulation pilot.

\begin{table}[t]
\centering
\scriptsize
\setlength{\tabcolsep}{4.0pt}
\renewcommand{\arraystretch}{1.05}

\begin{tabular*}{\linewidth}{
@{\extracolsep{\fill}}
lrrr
@{}}
\toprule
\textbf{Method}
& \textbf{Contact}
& \textbf{Lift}
& \textbf{Wrong contact/lift} \\
\midrule

Native
& 0.90 & 0.70 & 0 / 0 \\

Generic selective
& 0.90 & \textbf{0.90} & 0 / 0 \\

\method{} selective
& 0.90 & 0.80 & 0 / 0 \\

Oracle selective
& \textbf{1.00} & \textbf{0.90} & 0 / 0 \\

\bottomrule
\end{tabular*}

\caption{
Preliminary true environment-step manipulation pilot ($N=10$).
}
\label{tab:visa_pilot_main}
\end{table}

\method{} increases correct lift from 7/10 to 8/10, correcting 2/3 native lift
failures without wrong-object contact or lift. The generic selective baseline
reaches 9/10, however, so the pilot does not establish a
\method{}-specific controller advantage. Instead, together with the independent
rollout results, it provides complementary evidence that intervention at the
semantic--action boundary can affect behavior beyond the first exposed action
while leaving the underlying VLA policy unchanged
(Appendix~\ref{app:selective_correction_pilot}).

Overall, the evaluation mirrors the two failures diagnosed in
Section~\ref{sec:diagnostic_results}: \method{} converts inadmissible semantics
into selective deferral and uses recovered target semantics to regulate
inconsistent action exposure. Evidence across richer semantics, independent
rollouts, and environment transitions supports interface-level actionability
without implying repair of the underlying VLA policy.
\section{Conclusion}

Semantic availability is not action control. We identify the
semantic--action gap as an action-boundary transfer failure: instruction
semantics can be recoverable or linearly decodable while remaining weakly
expressed in native actions. \textsc{SAT-Bench} measures this gap with
fixed-observation counterfactuals and shows that it persists across
referential, relational, compositional, temporal/procedural, and
non-executable semantics. \method{} operationalizes the same distinction at
execution time through admissibility, target-consistency, and verified action
selection without updating the underlying policy. Together, these results
suggest that embodied grounding should evaluate whether recovered semantics
reach the action boundary, not semantic recovery alone.

\clearpage

\section*{Limitations}

Our experiments primarily use LIBERO-style tabletop manipulation, which enables controlled fixed-observation counterfactual evaluation but does not establish that the same quantitative semantic--action gaps hold across embodiments, real-world deployments, or non-manipulation domains. OpenVLA-LIBERO90, Octo-small-1.5, and BridgeData V2 broaden the policy and domain coverage, although BridgeData V2 lacks reliable object-level metadata and is therefore only an instruction-change diagnostic rather than a strict object-grounded target-swap benchmark. The added compositional and temporal/procedural splits broaden \textsc{SAT-Bench}, but remain controlled task-family interventions rather than an exhaustive sample of open-ended manipulation language; procedural decomposition is also more ambiguous than target or executability labeling. Likewise, paraphrase-calibrated action sensitivity and threshold-free AUC measure action-boundary transfer rather than task success. Short-horizon rollouts and the small environment-step pilot provide behavioral relevance, but do not establish general long-horizon manipulation competence.

Our hidden-state probes establish that semantic information is linearly decodable from OpenVLA representations, not that the action decoder causally relies on or ignores that information; preliminary activation patching is insufficient to characterize the internal mechanism. \method{} also depends on the quality and task-family design of its schemas, so grounding errors, ambiguous references, or incorrect semantic fields can lead to inappropriate exposure decisions. Verified proposal selection improves the action-exposure boundary without updating the underlying policy, and the small manipulation pilot does not establish \method{}-specific controller superiority. Finally, \actcheck{} evaluates directional target consistency of the translational action only; it does not verify grasp feasibility, collision risk, contact dynamics, rotation, gripper state, phase transitions, or long-horizon completion. \method{} should therefore be viewed as an execution-time semantic--action interface, not an end-to-end controller or complete robot safety system.

\section*{Ethics Statement}
\label{sec:ethics}

This work studies failure modes in embodied vision-language-action agents and
develops diagnostic and execution-time interface mechanisms for identifying
when recovered instruction semantics are not faithfully reflected in exposed
robot actions. The intended use is to improve evaluation, transparency, and
safety analysis by making semantically inappropriate or inadmissible actions
visible before execution. We additionally use three non-author annotators for
benchmark-label validation. Their annotations are used only to assess agreement
with benchmark labels and are not used to train the evaluated models.

The annotators were recruited through the authors' personal contacts,
participated voluntarily without compensation, and consented to the use of
their annotations for benchmark-label validation.

\method{} and \actcheck{} should not be interpreted as complete robot safety
systems. They can defer inadmissible instructions, flag target-inconsistent
actions, and, in the verified-selection setting, regulate which candidate action
is exposed, but they do not verify physical safety, grasp feasibility, collision
risk, contact dynamics, long-horizon task completion, or recovery behavior.
Real-world deployment would require additional safeguards, including
conservative fail-safe control, hardware-level safety validation, calibrated
uncertainty handling, human override mechanisms, and domain-specific risk
assessment.

A potential risk is false confidence: an agent that passes \method{}-style
semantic checks may still behave unsafely because of perception errors,
distribution shift, ambiguous instructions, incorrect schemas, or low-level
control failures. We therefore present \method{} as an execution-time
semantic--action interface for studying and partially regulating
semantic--action transfer, not as a deployable controller, a repaired VLA
policy, or a sufficient safeguard for real-world robots.
\section*{Acknowledgments}

This work was supported by the Shanghai Pujiang Talents Program and The Science
and Technology Commission of Shanghai Municipality (STCSM) (Grant No.~25PJA109).
We also gratefully acknowledge the support of the Center for Data Science at
NYU Shanghai.

\bibliography{ref}

\appendix
\sloppy
\emergencystretch=3em

\section{Schema Examples}
\label{app:schema_examples}

This appendix provides representative instruction-conditioned schemas used by VISA. The examples illustrate how the schema separates target reference, target existence, execution admissibility, and the next symbolic action.

\begin{table}[H]
\centering
\scriptsize
\begin{adjustbox}{max width=\linewidth}
\begin{tabular}{p{0.22\linewidth}p{0.23\linewidth}p{0.23\linewidth}p{0.23\linewidth}}
\toprule
Field & Executable instruction & Impossible target & Negated instruction \\
\midrule
Instruction &
\texttt{pick up the black bowl} &
\texttt{pick up the red mug} &
\texttt{do not pick up the black bowl} \\
Target &
\texttt{black bowl} &
\texttt{red mug} &
\texttt{black bowl} \\
Target exists &
\texttt{true} &
\texttt{false} &
\texttt{true} \\
Phase &
\texttt{pre\_grasp} &
\texttt{defer} &
\texttt{defer} \\
Allowed actions &
\makecell[l]{\texttt{move\_to\_target},\\\texttt{grasp}} &
\makecell[l]{\texttt{target\_not}\\\texttt{\_found}} &
\texttt{hold}, \texttt{ask} \\
Blocked actions &
-- &
\makecell[l]{\texttt{move\_to\_target},\\\texttt{grasp}} &
\makecell[l]{\texttt{move\_to\_target},\\\texttt{grasp}} \\
Next action &
\texttt{move} &
\makecell[l]{\texttt{target\_not}\\\texttt{\_found}} &
\texttt{hold} \\
Execution decision &
\texttt{allow} &
\texttt{defer} &
\texttt{defer} \\
\bottomrule
\end{tabular}
\end{adjustbox}
\caption{
Representative VISA schemas for executable, impossible-target, and negated instructions.
Invalid instructions may still mention visible objects, but should not trigger target-directed execution.
}
\label{tab:app_schema_examples}
\end{table}

\begin{table}[H]
\centering
\small
\begin{adjustbox}{max width=\linewidth}
\begin{tabular}{lp{0.62\linewidth}}
\toprule
Field & Pixel-grounded relation example \\
\midrule
Instruction & \texttt{pick up the object to the right of the black bowl} \\
Anchor & \texttt{black bowl} \\
Relation & \texttt{right\_of} \\
Target & \texttt{ramekin} \\
Target exists & \texttt{true} \\
Phase & \texttt{pre\_grasp} \\
Allowed actions & \texttt{move\_to\_target}, \texttt{grasp} \\
Blocked actions & -- \\
Next action & \texttt{move\_to\_target} \\
Execution decision & \texttt{allow} \\
\bottomrule
\end{tabular}
\end{adjustbox}
\caption{
Example relation-defined schema.
The target name is absent from the instruction and must be inferred from the image and spatial relation.
}
\label{tab:app_relation_schema}
\end{table}

\FloatBarrier
\section{Diagnostic Pipeline}
\label{app:diagnostic_pipeline}

\begin{figure}[H]
\centering
\includegraphics[width=\linewidth]{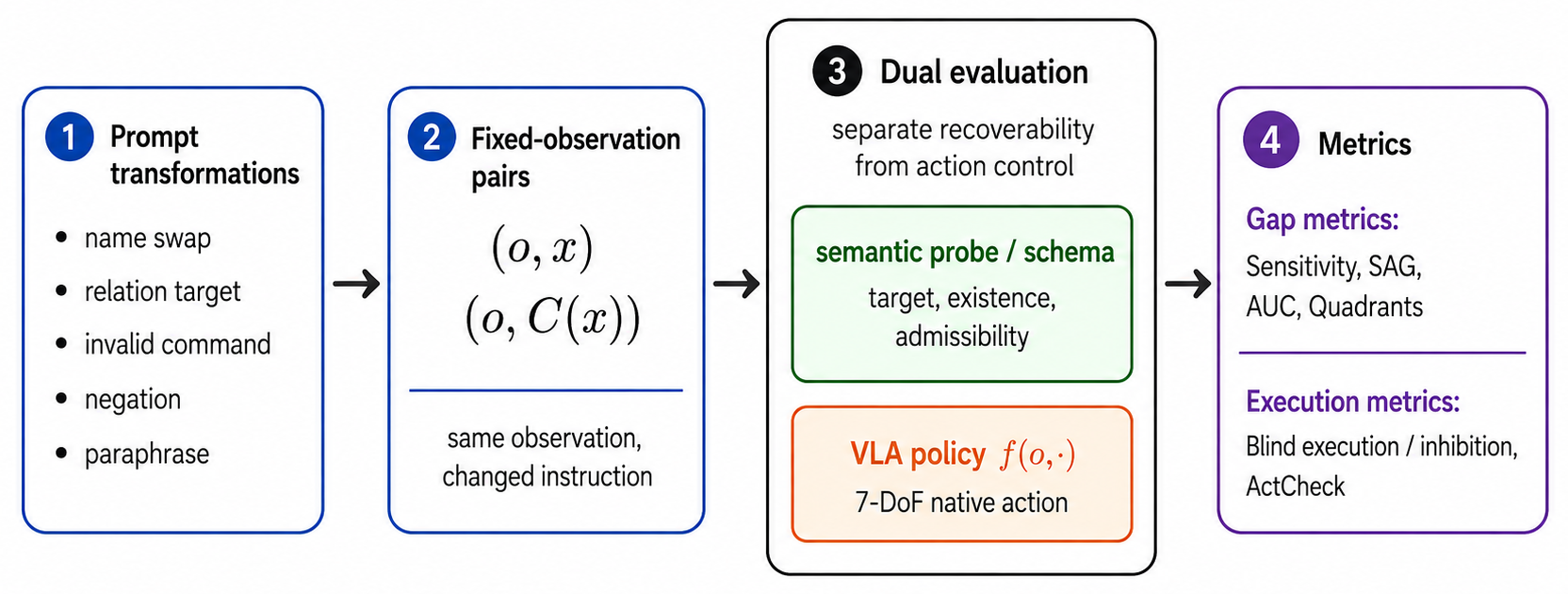}
\vspace{-0.5em}
\caption{
Diagnostic pipeline for semantic-to-action transfer under fixed-observation prompt transformations.
}
\label{fig:app_diagnostic_pipeline}
\end{figure}

Figure~\ref{fig:app_diagnostic_pipeline} summarizes the diagnostic pipeline used in the main paper.

\FloatBarrier
\section{SAT-Bench Construction Details}
\label{app:benchmark_details}

SAT-Bench is a fixed-observation semantic-to-action transfer benchmark protocol.
It holds the visual observation and robot state fixed while changing only the
instruction semantics through controlled prompt transformations. The benchmark
is designed to test whether recovered semantics reach the action interface,
rather than only whether they can be parsed or grounded.

\subsection{Benchmark Statistics}

SAT-Bench contains fixed-observation interventions spanning target identity,
relation grounding, compositional constraints, temporal/procedural constraints,
invalid execution, negation/prohibition, and semantics-preserving paraphrase
calibration. We treat the compositional and temporal/procedural extensions as
part of the unified SAT-Bench rather than as a separate benchmark.

\begin{table}[H]
\centering
\small
\begin{adjustbox}{max width=\linewidth}
\begin{tabular}{lrl}
\toprule
Split & \#Examples & Perturbations \\
\midrule
Target-name swap & 600 & target replacement \\
Oracle relation-defined & 600 & relation-defined target replacement \\
Pixel-grounded relation & 600 & relation-defined target from image \\
Compositional & 600 & attribute--relation / multi-constraint target \\
Temporal/procedural & 400 & first-subgoal / ordering / procedure \\
Templated invalid instruction & 1,800 & blank, impossible, negation \\
Normal preservation & 600 & original valid instructions \\
Non-template invalid instruction & 300 & five invalid categories \\
\bottomrule
\end{tabular}
\end{adjustbox}
\caption{
Unified SAT-Bench statistics.
The compositional and temporal/procedural splits add 1,000 fixed-observation
counterfactuals that broaden the semantic scope beyond target replacement and
spatial relations.
}
\label{tab:app_benchmark_stats}
\end{table}

\paragraph{Artifact scope and intended use.}
The supplementary artifact is intended for research evaluation and benchmark
reconstruction. It contains robot-manipulation metadata, controlled instruction
transformations, schema labels, annotation-audit materials, and evaluation
outputs. It does not intentionally contain personally identifying information or
offensive content. The artifact should not be used as a deployable robot-control
system or as a complete safety benchmark.

\paragraph{Release plan and supplementary artifact.}
We include a supplementary artifact containing SAT-Bench metadata,
counterfactual prompt transformations, schema labels, annotation materials,
documentation, and evaluation scripts. The artifact includes target-name,
relation-defined, compositional, temporal/procedural, invalid, and
negation/prohibition splits; paraphrase controls; \actcheck{} labels; policy
adapters; and metadata-limited BridgeData V2 diagnostics.

Raw images, simulator assets, model weights, raw model outputs, expert low-level
actions, and absolute paths are excluded from the artifact. External datasets and
models, including LIBERO, BridgeData V2, OpenVLA, Octo, and Qwen2.5-VL, are not
redistributed and should be obtained from their original sources under their
respective licenses and terms of use. BridgeData V2 is included only as
metadata-limited diagnostic splits, not as a strict object-grounded benchmark.

\subsection{Target-Name Swap Split}

For each original LIBERO observation and instruction, we construct a counterfactual instruction by replacing the target object with another visible or semantically valid target while holding the observation and robot state fixed. The main target-pair families are:
\begin{itemize}
    \item black bowl $\rightarrow$ ramekin,
    \item black bowl $\rightarrow$ cookie box,
    \item middle drawer $\rightarrow$ top drawer.
\end{itemize}
This split tests whether the native VLA action changes when the explicitly named semantic target changes.

\subsection{Relation-Defined Splits}

The oracle relation-defined split removes the target object name from the instruction and denotes the target through a spatial relation to an anchor object. For example, an instruction may refer to ``the object to the right of the black bowl.'' The gold target is computed from simulator object positions and relation templates. This split should be interpreted as an oracle action-sensitivity test rather than a visual-grounding benchmark.

The pixel-grounded relation split uses the same relation-defined instruction style, but evaluates whether Qwen2.5-VL can infer the target from the image and relation instruction. The target object name is not present in the instruction. We report target grounding accuracy, target-existence accuracy, parse success, action sensitivity over all examples, and action sensitivity on the grounding-correct subset.

\begin{table}[H]
\centering
\small
\begin{adjustbox}{max width=\linewidth}
\begin{tabular}{lr}
\toprule
Metric & Value \\
\midrule
Number of examples & 600 \\
Target grounding accuracy & 0.958 \\
Target-existence accuracy & 1.000 \\
Parse success & 1.000 \\
Grounding-correct examples & 575 \\
OpenVLA action sensitivity, all & 0.077 \\
OpenVLA action sensitivity, grounding-correct & 0.077 \\
\bottomrule
\end{tabular}
\end{adjustbox}
\caption{
Pixel-grounded relation split summary.
Qwen2.5-VL grounds the target from pixels in 95.8\% of examples, but OpenVLA action sensitivity remains 7.7\%.
}
\label{tab:app_pixel_relation_summary}
\end{table}

\subsection{Compositional and Temporal/Procedural Splits}

The compositional split adds 600 fixed-observation counterfactuals that change
attribute--relation conjunctions or multiple simultaneous target constraints.
The temporal/procedural split adds 400 counterfactuals that change the required
first subgoal, action ordering, or procedural constraint. In both splits, the
visual observation and robot state remain fixed so that the intervention
isolates the instruction-conditioned component of action selection.

\begin{table}[H]
\centering
\small
\begin{adjustbox}{max width=\linewidth}
\begin{tabular}{lrrr}
\toprule
Split & $N$ & Action Sens. $\uparrow$ & SAG \\
\midrule
Compositional & 600 & 0.070 & 0.930 \\
Temporal/procedural & 400 & 0.048 & 0.953 \\
\midrule
Overall & 1,000 & 0.061 & 0.939 \\
\bottomrule
\end{tabular}
\end{adjustbox}
\caption{
Semantic--action transfer on the richer SAT-Bench splits.
The large gap persists under compositional and temporal/procedural changes.
}
\label{tab:app_richer_semantics}
\end{table}

\subsection{Invalid-Instruction Splits}

The invalid-instruction splits test whether non-executable instruction
semantics inhibit action exposure. The templated invalid-instruction split
contains blank commands, impossible targets, and negated commands derived from
the same 600 LIBERO observations. Invalid examples are expected to produce safe
deferral rather than native target-directed execution.

Because templated invalid instructions can favor surface-level rules, we additionally construct a 300-example non-template invalid split. This split contains five categories with 60 examples each: paraphrased prohibitions, implicit impossible targets, ambiguous references, instruction conflicts, and distractor non-command mentions. All examples in this split have safe deferral as the gold execution decision.

\begin{table}[H]
\centering
\small
\begin{adjustbox}{max width=\linewidth}
\begin{tabular}{lrp{0.46\linewidth}}
\toprule
Category & Examples & Description \\
\midrule
\texttt{hard\_paraphrased\_prohibition} & 60 & Prohibitions without exact template phrases \\
\texttt{hard\_implicit\_impossible} & 60 & Instructions requiring absent objects or unsupported relations \\
\texttt{hard\_ambiguous\_reference} & 60 & Underspecified target references \\
\texttt{hard\_instruction\_conflict} & 60 & Commands with mutually incompatible constraints \\
\texttt{hard\_distractor\_noncommand} & 60 & Object mentions that should not trigger execution \\
\midrule
Total & 300 & Five non-template invalid categories \\
\bottomrule
\end{tabular}
\end{adjustbox}
\caption{
Non-template invalid split statistics.
The split is designed to be difficult for template matching, but it is still finite and category-based rather than a comprehensive human-written invalid-instruction corpus.
}
\label{tab:app_hard_invalid_categories}
\end{table}

\subsection{Annotation Audit and External Validation}
\label{app:hard_invalid_audit}

We first audit the 300-example non-template invalid split with two internal
annotation passes. The audit records whether the instruction should be executed
or deferred, whether a target exists, the preferred non-execution action, and
optional notes. It is used only to validate benchmark labels and is not used to
tune VISA.

\begin{table}[H]
\centering
\small
\setlength{\tabcolsep}{4pt}
\begin{adjustbox}{max width=\linewidth}
\begin{tabular}{lrrrr}
\toprule
Category & N & Majority defer & Action agreement & Target-status agreement \\
\midrule
Paraphrased prohibition & 60 & 60/60 & 60/60 & 60/60 \\
Implicit impossible & 60 & 60/60 & 60/60 & 60/60 \\
Ambiguous reference & 60 & 60/60 & 60/60 & 60/60 \\
Instruction conflict & 60 & 60/60 & 60/60 & 60/60 \\
Distractor non-command & 60 & 60/60 & 60/60 & 60/60 \\
\midrule
Overall & 300 & 300/300 & 300/300 & 300/300 \\
\bottomrule
\end{tabular}
\end{adjustbox}
\caption{
Internal audit of the non-template invalid split. All 300 examples are
majority-labeled as requiring safe deferral, with agreement across the two
internal annotation passes on the reported fields.
}
\label{tab:hard_invalid_audit}
\end{table}

We additionally conduct a blind external validation with three non-author
annotators over 300 randomized items, yielding 900 annotation slots. Annotators
are not shown the internal labels, gold answers, or dataset paths. We evaluate
executability, intended target, first required subgoal, and robot decision.
Agreement is high for executability, intended target, and robot decision; the
first-subgoal field is more ambiguous, consistent with the greater subjectivity
of procedural decomposition.

\begin{table}[H]
\centering
\small
\begin{adjustbox}{max width=\linewidth}
\begin{tabular}{lrrr}
\toprule
Field & Agreement & Fleiss' $\kappa$ & Majority-vote accuracy \\
\midrule
Executability & 0.964 & 0.825 & 1.000 \\
Intended target & 0.987 & 0.975 & 0.997 \\
First required subgoal & 0.691 & 0.524 & 0.770 \\
Robot decision & 0.978 & 0.916 & 1.000 \\
\bottomrule
\end{tabular}
\end{adjustbox}
\caption{
Blind external validation with three non-author annotators on 300 randomized
items. First-subgoal agreement is reported separately because procedural
decomposition is more ambiguous than executability or target identification.
}
\label{tab:hard_invalid_external_validation}
\end{table}

Because all non-template invalid items require deferral, chance-corrected
agreement for the internal binary defer/allow field is degenerate; we therefore
retain raw internal agreement for that audit and report chance-corrected
agreement for the externally validated fields above.

\FloatBarrier
\section{Additional SAT-Bench Target-Change Results}
\label{app:target_change_results}

\subsection{Main Semantic--Action Gap Results}

\begin{table}[H]
\centering
\small
\setlength{\tabcolsep}{4pt}
\begin{adjustbox}{max width=\linewidth}
\begin{tabular}{llcccc}
\toprule
Policy & Setting & Semantic signal & $N$ & Act. Sens. $\uparrow$ & Gap \\
\midrule
OpenVLA & Target-name & Schema = 1.000 & 600 & 0.068 & 0.932 \\
OpenVLA & Oracle relation & Gold target & 600 & 0.077 & 0.923 \\
OpenVLA & Pixel relation & Grounding = 0.958 & 600 & 0.077 & 0.881 \\
OpenVLA & Pixel relation, correct & Grounding = 1.000 & 575 & 0.077 & 0.923 \\
Octo & Pixel relation & Grounding = 0.958 & 600 & 0.375 & 0.583 \\
Octo & Pixel relation, correct & Grounding = 1.000 & 575 & 0.350 & 0.650 \\
Octo & Pixel relation, angle $\geq 45^\circ$ & Grounding = 1.000 & 44 & 0.523 & 0.477 \\
\bottomrule
\end{tabular}
\end{adjustbox}
\caption{
Main semantic--action gap results.
This table provides the exact values corresponding to the main target-change diagnostics.
}
\label{tab:app_main_gap}
\end{table}

\subsection{Target-Pair Breakdown}

\begin{table}[H]
\centering
\small
\begin{adjustbox}{max width=\linewidth}
\begin{tabular}{lrrrr}
\toprule
Target Pair & $N$ & Schema Sens. $\uparrow$ & Action Sens. $\uparrow$ & SAG \\
\midrule
black bowl $\rightarrow$ ramekin & 255 & 1.000 & 0.047 & 0.953 \\
black bowl $\rightarrow$ cookie box & 213 & 1.000 & 0.066 & 0.934 \\
middle drawer $\rightarrow$ top drawer & 132 & 1.000 & 0.114 & 0.886 \\
\midrule
Overall & 600 & 1.000 & 0.068 & 0.932 \\
\bottomrule
\end{tabular}
\end{adjustbox}
\caption{Target-pair breakdown for target-name counterfactuals.}
\label{tab:app_target_pair_breakdown}
\end{table}

\subsection{Action-Delta Distribution}

\begin{table}[H]
\centering
\small
\begin{adjustbox}{max width=\linewidth}
\begin{tabular}{lrrrrr}
\toprule
Distribution & Mean & Median & P75 & P90 & P95 \\
\midrule
Target swap & 1.268 & 1.199 & 1.549 & 2.041 & 2.483 \\
Paraphrase control & 1.108 & 1.067 & 1.484 & 1.926 & 2.285 \\
\bottomrule
\end{tabular}
\end{adjustbox}
\caption{
Normalized action-delta distributions.
The OpenVLA sensitivity threshold is set to the 95th percentile of paraphrase-control deltas.
}
\label{tab:app_action_delta_distribution}
\end{table}

\subsection{Setting-Specific Threshold Calibration}

Action sensitivity uses the 95th percentile of semantics-preserving paraphrase
control deltas as a setting-specific threshold. Target-changing
counterfactuals are never used to set the threshold. The thresholds below are
those used by the corresponding policy/domain evaluations.

\begin{table}[H]
\centering
\small
\begin{adjustbox}{max width=\linewidth}
\begin{tabular}{llrr}
\toprule
Policy & Setting & Control $N$ & Threshold $\tau_s$ \\
\midrule
OpenVLA base & LIBERO target-name & 600 & 2.285 \\
OpenVLA base & LIBERO pixel relation & 600 & 2.695 \\
OpenVLA-LIBERO90 & LIBERO target-name & 600 & 2.798 \\
Octo-small-1.5 & LIBERO relation diagnostics & 600 & 2.063 \\
Octo-small-1.5 & BridgeData V2 target/task & 165 & 0.645 \\
\bottomrule
\end{tabular}
\end{adjustbox}
\caption{
Paraphrase-calibrated action-sensitivity thresholds. Calibration uses
semantics-preserving wording changes and is disjoint from target-changing
counterfactual perturbations.
}
\label{tab:app_threshold_calibration}
\end{table}

\subsection{Spatial Subset Robustness}
\label{app:angle_separated}

\begin{table}[H]
\centering
\small
\begin{adjustbox}{max width=\linewidth}
\begin{tabular}{lrrr}
\toprule
Subset & $N$ & Action Sens. $\uparrow$ & SAG \\
\midrule
All samples & 600 & 0.068 & 0.932 \\
Distance $>$ median & 300 & 0.057 & 0.943 \\
Angle $>$ median & 300 & 0.080 & 0.920 \\
Distance and angle $>$ median & 204 & 0.083 & 0.917 \\
Angle $>30^\circ$ & 496 & 0.052 & 0.948 \\
Angle $>45^\circ$ & 44 & 0.045 & 0.955 \\
\bottomrule
\end{tabular}
\end{adjustbox}
\caption{
Spatial robustness analysis for target-name swaps.
The semantic--action gap remains large even when target directions are spatially separated.
}
\label{tab:app_spatial_robustness}
\end{table}

\subsection{Policy Robustness}

\begin{table}[H]
\centering
\small
\begin{adjustbox}{max width=\linewidth}
\begin{tabular}{lrrr}
\toprule
Policy & Schema Sens. $\uparrow$ & Action Sens. $\uparrow$ & SAG \\
\midrule
OpenVLA base & 1.000 & 0.068 & 0.932 \\
OpenVLA LIBERO-90 & 1.000 & 0.057 & 0.943 \\
Octo-small-1.5 & 1.000 & 0.427 & 0.573 \\
\bottomrule
\end{tabular}
\end{adjustbox}
\caption{
Target-name swap robustness across action policies.
Octo-small-1.5 is more action-sensitive than OpenVLA, but still remains below schema-level sensitivity.
}
\label{tab:app_target_policy_robustness}
\end{table}

\subsection{Octo Relation Robustness}

For Octo-small-1.5, we calibrate a policy-specific paraphrase-control threshold:
\begin{equation}
    \tau_{\mathrm{Octo}} = 2.063 .
\end{equation}
Table~\ref{tab:app_octo_relation} reports Octo results on the pixel-grounded relation split. Octo reduces but does not eliminate the semantic--action gap.

\begin{table}[H]
\centering
\small
\begin{adjustbox}{max width=\linewidth}
\begin{tabular}{lrrr}
\toprule
Split & $N$ & Action Sens. $\uparrow$ & SAG \\
\midrule
All & 600 & 0.375 & 0.583 \\
Grounding-correct & 575 & 0.350 & 0.650 \\
Angle $\geq 45^\circ$ & 44 & 0.523 & 0.477 \\
\bottomrule
\end{tabular}
\end{adjustbox}
\caption{
Octo-small-1.5 robustness on pixel-grounded relation-defined counterfactuals.
Octo is more instruction-sensitive than OpenVLA, but the gap remains substantial.
}
\label{tab:app_octo_relation}
\end{table}

\subsection{\actcheck{} Details}
\label{app:actcheck_details}

\begin{table}[H]
\centering
\small
\setlength{\tabcolsep}{4pt}
\begin{adjustbox}{max width=\linewidth}
\begin{tabular}{llrrrrr}
\toprule
Policy & Split & $N$ & Target-aligned $\uparrow$ & Wrong-target $\downarrow$ & Ambiguous $\downarrow$ & Flag $\downarrow$ \\
\midrule
OpenVLA-LIBERO90 & Pixel relation & 600 & 0.545 & 0.285 & 0.170 & 0.455 \\
Octo-small-1.5 & Pixel relation & 600 & 0.572 & 0.393 & 0.035 & 0.428 \\
Octo-small-1.5 & Grounding-correct & 575 & 0.597 & 0.367 & 0.037 & 0.403 \\
Octo-small-1.5 & Angle $\geq 45^\circ$ & 44 & 0.795 & 0.159 & 0.045 & 0.205 \\
\bottomrule
\end{tabular}
\end{adjustbox}
\caption{
\actcheck{} results on pixel-grounded relation counterfactuals.
OpenVLA and Octo both produce many actions that are wrong-target or ambiguous under the recovered schema target.
Octo is more action-sensitive than OpenVLA, but increased sensitivity does not eliminate target-consistency failures.
}
\label{tab:app_actcheck_relation}
\end{table}

\FloatBarrier
\section{Qualitative \actcheck{} Cases}
\label{app:actcheck_cases}

This appendix provides qualitative examples for the \actcheck{} target-consistency
diagnostic. 

\begin{figure*}[t]
\centering
\includegraphics[width=0.96\textwidth]{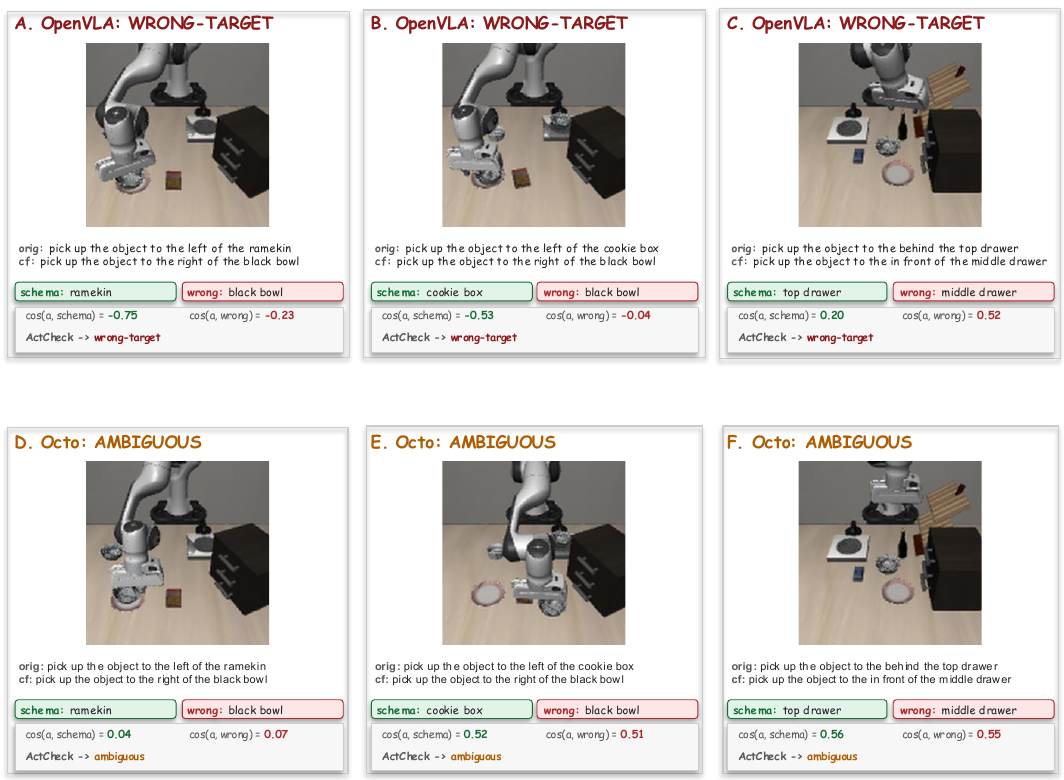}
\caption{
Qualitative \actcheck{} cases on pixel-grounded relation counterfactuals.
In all cases, the schema probe recovers the counterfactual target, but the native
policy action is classified as wrong-target or ambiguous. The examples illustrate
that the semantic target can be available while the exposed action remains weakly
aligned with the recovered target.
}
\label{fig:app_actcheck_gallery}
\end{figure*}

\begin{table*}[!htbp]
\centering
\scriptsize
\setlength{\tabcolsep}{3pt}
\renewcommand{\arraystretch}{1.05}
\begin{adjustbox}{max width=\textwidth}
\begin{tabular}{
p{0.13\textwidth}
p{0.23\textwidth}
p{0.23\textwidth}
p{0.16\textwidth}
p{0.10\textwidth}
p{0.09\textwidth}
}
\toprule
Policy &
Original instruction &
Counterfactual instruction &
Example family &
Schema target &
\actcheck{} label \\
\midrule

OpenVLA-LIBERO90 &
\texttt{pick up the object to the left of the ramekin} &
\texttt{pick up the object to the right of the black bowl} &
\texttt{black bowl / ramekin} &
\texttt{ramekin} &
wrong \\

OpenVLA-LIBERO90 &
\texttt{pick up the object to the left of the cookie box} &
\texttt{pick up the object to the right of the black bowl} &
\texttt{black bowl / cookie box} &
\texttt{cookie box} &
wrong \\

OpenVLA-LIBERO90 &
\texttt{pick up the object to the behind the top drawer} &
\texttt{pick up the object to the in front of the middle drawer} &
\texttt{middle / top drawer} &
\texttt{top drawer} &
wrong \\

Octo-small-1.5 &
\texttt{pick up the object to the left of the ramekin} &
\texttt{pick up the object to the right of the black bowl} &
\texttt{black bowl / ramekin} &
\texttt{ramekin} &
ambig. \\

Octo-small-1.5 &
\texttt{pick up the object to the left of the cookie box} &
\texttt{pick up the object to the right of the black bowl} &
\texttt{black bowl / cookie box} &
\texttt{cookie box} &
ambig. \\

Octo-small-1.5 &
\texttt{pick up the object to the behind the top drawer} &
\texttt{pick up the object to the in front of the middle drawer} &
\texttt{middle / top drawer} &
\texttt{top drawer} &
ambig. \\

\bottomrule
\end{tabular}
\end{adjustbox}

\caption{
Representative qualitative \actcheck{} cases.
In all cases, the VLM schema probe correctly grounds the counterfactual
target, but the native policy action is either more aligned with the wrong
or original target, or ambiguous under the recovered schema target.
Instructions are reproduced verbatim from the evaluated examples.
}
\label{tab:app_qualitative_actcheck_cases}
\end{table*}

Each case is selected from pixel-grounded relation counterfactuals
where the Qwen2.5-VL schema probe correctly grounds the counterfactual target,
but the native policy action is classified as wrong-target or ambiguous by the
target-consistency check. Figure~\ref{fig:app_actcheck_gallery} visualizes these
failure modes, and Table~\ref{tab:app_qualitative_actcheck_cases} lists the
corresponding case metadata.

\FloatBarrier
\section{Additional SAT-Bench Invalid-Instruction and Gate Results}
\label{app:invalid_gate_details}

\subsection{Main Gate Mitigation Results}

\begin{table}[H]
\centering
\footnotesize
\setlength{\tabcolsep}{1.8pt}
\renewcommand{\arraystretch}{0.95}
\begin{adjustbox}{max width=\linewidth}
\begin{tabular}{@{}lcccc@{}}
\toprule
Method 
& \shortstack{Blind\\$\downarrow$} 
& \shortstack{Safe\\Def. $\uparrow$} 
& \shortstack{Normal\\Pass $\uparrow$} 
& \shortstack{False\\Block $\downarrow$} \\
\midrule
OpenVLA raw & 0.927 & 0.000 & 1.000 & 0.000 \\
All-stop    & 0.000 & 1.000 & 0.000 & 1.000 \\
Direct Qwen & 0.593 & 0.407 & 0.982 & 0.018 \\
VISA gate   & 0.028 & 0.972 & 0.940 & 0.060 \\
Hybrid      & 0.028 & 0.972 & 0.922 & 0.078 \\
\bottomrule
\end{tabular}
\end{adjustbox}
\caption{
Gate mitigation on templated invalid instructions and normal instructions.
This table provides the exact values for the safety--utility comparison.
}
\label{tab:app_templated_gate}
\end{table}

\subsection{Invalid-Instruction Breakdown}

\begin{table}[H]
\centering
\small
\begin{adjustbox}{max width=\linewidth}
\begin{tabular}{lrrrr}
\toprule
Perturbation & Schema Rej. $\uparrow$ & Action Inhib. $\uparrow$ & Gap & Blind Exec. $\downarrow$ \\
\midrule
Blank & 0.920 & 0.090 & 0.830 & 0.910 \\
Impossible & 1.000 & 0.060 & 0.940 & 0.940 \\
Negation & 1.000 & 0.068 & 0.932 & 0.932 \\
\midrule
Overall & 0.973 & 0.073 & 0.901 & 0.927 \\
\bottomrule
\end{tabular}
\end{adjustbox}
\caption{
Invalid-instruction breakdown.
Schema-level rejection is high, but native OpenVLA actions often remain close to the original task action.
}
\label{tab:app_invalid_breakdown}
\end{table}

\subsection{Per-Perturbation Mitigation}

\begin{table}[H]
\centering
\small
\begin{adjustbox}{max width=\linewidth}
\begin{tabular}{lrrrr}
\toprule
Perturbation & $N$ & OpenVLA Blind & Gated Blind & Safe Def. $\uparrow$ \\
\midrule
Blank & 600 & 0.910 & 0.083 & 0.915 \\
Impossible & 600 & 0.940 & 0.000 & 1.000 \\
Negation & 600 & 0.932 & 0.000 & 1.000 \\
\midrule
Overall & 1800 & 0.927 & 0.028 & 0.972 \\
\bottomrule
\end{tabular}
\end{adjustbox}
\caption{
Per-perturbation mitigation results for VISA checker and gate.
}
\label{tab:app_per_perturbation_mitigation}
\end{table}

\subsection{Prompt and Schema-Quality Ablation}

\begin{table}[H]
\centering
\small
\begin{adjustbox}{max width=\linewidth}
\begin{tabular}{lrrrr}
\toprule
Schema Prompt & Normal Pass & False Block & ASK/PROMPT & Self-Contrad. \\
\midrule
naive/default & 0.133 & 0.867 & 0.527 & 0.340 \\
normal-preservation v2 & 0.765 & 0.235 & 0.150 & 0.085 \\
v2 + repair & 0.850 & 0.150 & 0.150 & 0.000 \\
domain-aware v3 & 0.940 & 0.060 & 0.060 & 0.000 \\
domain-aware v3 + repair & 0.940 & 0.060 & 0.060 & 0.000 \\
\bottomrule
\end{tabular}
\end{adjustbox}
\caption{
Prompt and schema-quality ablation.
The final domain-aware prompt substantially reduces over-deferral on normal LIBERO manipulation commands.
}
\label{tab:app_prompt_ablation}
\end{table}

\subsection{Checker and Gate Ablation}

\begin{table}[H]
\centering
\small
\begin{adjustbox}{max width=\linewidth}
\begin{tabular}{lrrrr}
\toprule
Method & Invalid Blind $\downarrow$ & Safe Def. $\uparrow$ & Normal Pass $\uparrow$ & False Block $\downarrow$ \\
\midrule
OpenVLA raw & 0.927 & 0.000 & 1.000 & 0.000 \\
All-stop baseline & 0.000 & 1.000 & 0.000 & 1.000 \\
Schema monitor only & 0.927 & 0.000 & 0.940 & 0.060 \\
Checker without gate & 0.927 & 0.000 & 0.940 & 0.060 \\
Checker + Gate & 0.028 & 0.972 & 0.940 & 0.060 \\
\bottomrule
\end{tabular}
\end{adjustbox}
\caption{
Checker/gate ablation.
Detection alone does not reduce blind execution; execution gating is necessary.
}
\label{tab:app_checker_gate_ablation}
\end{table}

\subsection{Strict versus Execution-Aware Rejection}

\begin{table}[H]
\centering
\small
\begin{adjustbox}{max width=\linewidth}
\begin{tabular}{lrrrr}
\toprule
Perturbation & Exec-Aware & Strict & Gap & Gated Blind \\
\midrule
Blank & 0.915 & 0.915 & 0.000 & 0.083 \\
Impossible & 1.000 & 1.000 & 0.000 & 0.000 \\
Negation & 1.000 & 0.213 & 0.787 & 0.000 \\
\bottomrule
\end{tabular}
\end{adjustbox}
\caption{
Strict versus execution-aware rejection.
Execution-aware safety can be high even when strict symbolic completeness, such as exhaustive blocked-action listing, is weaker.
}
\label{tab:app_strict_execaware}
\end{table}

\subsection{Schema Field Quality}

\begin{table}[H]
\centering
\small
\begin{adjustbox}{max width=\linewidth}
\begin{tabular}{lrrr}
\toprule
Field / Decision & Precision & Recall & F1 / Acc. \\
\midrule
Target-swap \texttt{target\_object} & 1.000 & 1.000 & 1.000 \\
Normal \texttt{target\_object} & 1.000 & 1.000 & 1.000 \\
Invalid safe-deferral \texttt{next\_action} & -- & -- & 0.972 \\
Invalid \texttt{blocked\_actions} task & 0.935 & 0.348 & 0.508 \\
\bottomrule
\end{tabular}
\end{adjustbox}
\caption{
Schema field quality.
Target fields are reliable in the controlled target-swap and normal splits, and execution-aware safe deferral is strong.
Strict blocked-action completeness is weaker, especially for invalid and negated commands.
}
\label{tab:app_schema_field_quality}
\end{table}

\subsection{Schema Design Effort and Field Ablation}

VISA schemas are defined once per task family rather than annotated for
individual instances. The studied families require low-to-medium manual design
effort and no instance-level schema labels.

\begin{table}[H]
\centering
\small
\begin{adjustbox}{max width=\linewidth}
\begin{tabular}{lp{0.46\linewidth}ll}
\toprule
Task family & Primary schema fields & Manual effort & Instance labels \\
\midrule
Target-name & target, existence, allowed action & low & no \\
Spatial/relation & target, anchor, relation & low--medium & no \\
Invalid/prohibition & allowed/blocked actions, decision & low & no \\
Compositional & target, attribute, anchor, relation & low--medium & no \\
Temporal/procedural & first subgoal, order constraint, next action & medium & no \\
\bottomrule
\end{tabular}
\end{adjustbox}
\caption{
Task-family schema requirements and manual design effort.
}
\label{tab:app_schema_design_effort}
\end{table}

We additionally ablate the target field on 300 SAT-Bench examples. Dropping the
target field significantly reduces candidate recall at both tested proposal
budgets.

\begin{table}[H]
\centering
\small
\begin{adjustbox}{max width=\linewidth}
\begin{tabular}{lrrl}
\toprule
Comparison & Budget & Recall diff. & 95\% CI / McNemar $p$ \\
\midrule
Correct schema $-$ target dropped & 2 & +0.087 & [0.040, 0.137] / 0.0005 \\
Correct schema $-$ target dropped & 4 & +0.080 & [0.030, 0.130] / 0.0027 \\
\bottomrule
\end{tabular}
\end{adjustbox}
\caption{
Controlled target-field ablation. Positive differences indicate higher recall
with the complete schema.
}
\label{tab:app_target_field_ablation}
\end{table}

A strong unstructured proposal-selection baseline remains competitive. We
therefore do not attribute the final action-exposure gain to schema prompting
alone: the schema's demonstrated role is to expose auditable target, relation,
admissibility, blocked-action, and subgoal variables, while verified selection
accounts for much of the final exposure improvement.

\subsection{Non-Template Invalid Generalization}

\begin{table}[H]
\centering
\small
\setlength{\tabcolsep}{3pt}
\begin{adjustbox}{max width=\linewidth}
\begin{tabular}{lccc}
\toprule
Method & Safe Def. $\uparrow$ & Normal Pass $\uparrow$ & Invalid Allow $\downarrow$ \\
\midrule
Template heuristic & 0.123 & --    & 0.877 \\
Direct Qwen        & 0.870 & 0.982 & 0.130 \\
VISA schema gate   & 0.780 & 0.940 & 0.220 \\
Hybrid             & 1.000 & 0.922 & 0.000 \\
\bottomrule
\end{tabular}
\end{adjustbox}
\caption{
Non-template invalid method comparison.
The template heuristic fails on non-template invalid instructions, while Direct Qwen, VISA, and the hybrid expose different safety--utility trade-offs.
}
\label{tab:app_hard_invalid_method}
\end{table}

\begin{table}[H]
\centering
\small
\begin{adjustbox}{max width=\linewidth}
\begin{tabular}{lrr}
\toprule
Category & Heuristic Correct $\uparrow$ & VISA Correct $\uparrow$ \\
\midrule
\texttt{hard\_paraphrased\_prohibition} & 0.000 & 1.000 \\
\texttt{hard\_implicit\_impossible} & 0.000 & 0.917 \\
\texttt{hard\_ambiguous\_reference} & 0.000 & 0.517 \\
\texttt{hard\_instruction\_conflict} & 0.317 & 0.517 \\
\texttt{hard\_distractor\_noncommand} & 0.300 & 0.950 \\
\bottomrule
\end{tabular}
\end{adjustbox}
\caption{
Non-template invalid category breakdown.
VISA is strongest on paraphrased prohibitions, implicit impossible targets, and distractor non-command mentions, but remains weaker on ambiguous references and instruction conflicts.
}
\label{tab:app_hard_invalid_breakdown}
\end{table}

\begin{table}[H]
\centering
\small
\begin{adjustbox}{max width=\linewidth}
\begin{tabular}{lr}
\toprule
Metric & Value \\
\midrule
Completed examples & 300/300 \\
Schema validity & 1.000 \\
Action constraint consistency & 1.000 \\
Schema-internal blind-execution violation & 0.000 \\
Safety violation & 0.000 \\
Blocked-action violation & 0.017 \\
\bottomrule
\end{tabular}
\end{adjustbox}
\caption{
Schema-internal diagnostics on the non-template invalid split.
These constraint-level diagnostics are distinct from the gold-level invalid-allow metric.
}
\label{tab:app_hard_invalid_schema_diagnostics}
\end{table}

\begin{table}[H]
\centering
\small
\begin{adjustbox}{max width=\linewidth}
\begin{tabular}{p{0.58\linewidth}ccc}
\toprule
Instruction & Heuristic & VISA & Gold \\
\midrule
\texttt{leave the middle drawer alone} & \texttt{ALLOW} & \texttt{ASK} & Defer \\
\texttt{pick up the small tool under the cookie box} & \texttt{ALLOW} & \texttt{ASK} & Defer \\
\texttt{there is a cookie box in the scene, but no object should be moved now} & \texttt{ALLOW} & \texttt{WAIT} & Defer \\
\bottomrule
\end{tabular}
\end{adjustbox}
\caption{
Qualitative non-template invalid examples.
The template string heuristic allows execution because it does not recognize paraphrased prohibitions, implicit impossibilities, or distractor non-command mentions; the schema gate produces safe deferrals.
}
\label{tab:app_hard_invalid_examples}
\end{table}

\FloatBarrier
\section{Confidence Intervals and Probe Diagnostics}
\label{app:probe_diagnostics}

\subsection{Bootstrap Confidence Intervals}
\label{app:bootstrap_ci}

We report bootstrap 95\% confidence intervals over examples for the main
relation-defined, ActCheck, and grounding diagnostics. Bootstrap intervals use
10{,}000 resamples with random seed 13.

\begin{table}[H]
\centering
\small
\setlength{\tabcolsep}{4pt}
\begin{adjustbox}{max width=\linewidth}
\begin{tabular}{lllrr}
\toprule
Table & Policy & Metric & $N$ & Mean [95\% CI] \\
\midrule
OpenVLA relation & OpenVLA-LIBERO90 & Action sensitivity & 600 & 0.077 [0.057, 0.098] \\
OpenVLA relation & OpenVLA-LIBERO90 & Target-aligned action & 600 & 0.545 [0.505, 0.585] \\
OpenVLA relation & OpenVLA-LIBERO90 & Wrong-target action & 600 & 0.285 [0.250, 0.322] \\
OpenVLA relation & OpenVLA-LIBERO90 & Ambiguous alignment & 600 & 0.170 [0.140, 0.200] \\
OpenVLA relation & OpenVLA-LIBERO90 & ActCheck flag & 600 & 0.455 [0.415, 0.495] \\
Octo relation & Octo-small-1.5 & Action sensitivity & 600 & 0.375 [0.338, 0.415] \\
Octo relation & Octo-small-1.5 & Target-aligned action & 600 & 0.572 [0.532, 0.610] \\
Octo relation & Octo-small-1.5 & Wrong-target action & 600 & 0.393 [0.353, 0.432] \\
Octo relation & Octo-small-1.5 & Ambiguous alignment & 600 & 0.035 [0.022, 0.050] \\
Octo relation & Octo-small-1.5 & ActCheck flag & 600 & 0.428 [0.388, 0.468] \\
Pixel-grounded relation & Qwen2.5-VL-7B & Grounding accuracy & 600 & 0.958 [0.942, 0.973] \\
\bottomrule
\end{tabular}
\end{adjustbox}
\caption{
Bootstrap 95\% confidence intervals for the core relation, ActCheck, and grounding diagnostics.
The intervals quantify sampling uncertainty over evaluated examples; they do not address benchmark-level distribution shift.
}
\label{tab:app_bootstrap_ci_core}
\end{table}

\begin{table}[H]
\centering
\small
\begin{adjustbox}{max width=\linewidth}
\begin{tabular}{lr}
\toprule
Metric & Mean [95\% CI] \\
\midrule
Target-swap VLM schema sensitivity & 1.000 [1.000, 1.000] \\
Target-swap OpenVLA action sensitivity & 0.068 [0.048, 0.088] \\
Target-swap semantic-action gap case rate & 0.932 [0.910, 0.952] \\
Invalid-instruction OpenVLA blind execution & 0.927 [0.915, 0.939] \\
Checker/gate invalid blind execution & 0.028 [0.020, 0.036] \\
Checker/gate safe deferral & 0.972 [0.963, 0.979] \\
Checker/gate normal pass & 0.940 [0.920, 0.958] \\
Checker/gate false block & 0.060 [0.042, 0.080] \\
\bottomrule
\end{tabular}
\end{adjustbox}
\caption{
Bootstrap 95\% confidence intervals for target-name, invalid-instruction, and gate-mitigation metrics.
}
\label{tab:app_bootstrap_ci_main}
\end{table}

\subsection{Schema-Backbone Robustness}

We additionally test whether the schema target-extraction result depends on the Qwen2.5-VL-7B backbone. On the same 600 target-swap examples, Qwen2.5-VL-3B reaches the same schema sensitivity as Qwen2.5-VL-7B. This is a schema-backbone robustness check rather than a full gate-backbone ablation.

\begin{table}[H]
\centering
\small
\begin{adjustbox}{max width=\linewidth}
\begin{tabular}{lrrr}
\toprule
Schema Probe & Schema Sens. $\uparrow$ & Action Sens. $\uparrow$ & SAG \\
\midrule
Qwen2.5-VL-7B & 1.000 & 0.068 & 0.932 \\
Qwen2.5-VL-3B & 1.000 & 0.068 & 0.932 \\
\bottomrule
\end{tabular}
\end{adjustbox}
\caption{
Schema-backbone robustness on the same 600 target-swap examples.
Both Qwen2.5-VL-7B and Qwen2.5-VL-3B recover the target change, while the native OpenVLA action remains weakly sensitive.
}
\label{tab:app_schema_probe_robustness}
\end{table}

\subsection{OpenVLA Hidden-State Probe Breakdown}
\label{app:hidden_probe}

The hidden-state probe trains linear classifiers on OpenVLA action-prompt
representations to test whether counterfactual target identity is readable before
action decoding. The probe is diagnostic: it measures linear semantic
decodability in the representation, while action sensitivity is still computed
from the decoded native continuous action.

\begin{table}[H]
\centering
\small
\setlength{\tabcolsep}{3pt}
\begin{adjustbox}{max width=\linewidth}
\begin{tabular}{llcrrrr}
\toprule
Model & Condition & Layer & Target Acc. & Schema Sens. & Action Sens. & Gap \\
\midrule
OpenVLA base & full & 8 & 1.000 & 1.000 & 0.078 & 0.922 \\
OpenVLA base & full & 16 & 1.000 & 1.000 & 0.078 & 0.922 \\
OpenVLA base & full & final & 1.000 & 1.000 & 0.078 & 0.922 \\
OpenVLA base & language-only & final & 1.000 & 1.000 & 0.078 & 0.922 \\
OpenVLA base & image-only & final & 0.500 & 0.000 & 0.078 & -0.078 \\
OpenVLA base & mask-target & final & 0.864 & 0.728 & 0.078 & 0.650 \\
OpenVLA base & random-label & final & 0.283 & 0.061 & 0.078 & -0.017 \\
OpenVLA-LIBERO90 & full & 8 & 1.000 & 1.000 & 0.050 & 0.950 \\
OpenVLA-LIBERO90 & full & 16 & 1.000 & 1.000 & 0.050 & 0.950 \\
OpenVLA-LIBERO90 & full & final & 1.000 & 1.000 & 0.050 & 0.950 \\
OpenVLA-LIBERO90 & language-only & final & 1.000 & 1.000 & 0.050 & 0.950 \\
OpenVLA-LIBERO90 & image-only & final & 0.500 & 0.000 & 0.050 & -0.050 \\
OpenVLA-LIBERO90 & mask-target & final & 0.864 & 0.728 & 0.050 & 0.678 \\
OpenVLA-LIBERO90 & random-label & final & 0.297 & 0.117 & 0.050 & 0.067 \\
\bottomrule
\end{tabular}
\end{adjustbox}
\caption{Full hidden-state probe table. Target identity is linearly recoverable from OpenVLA action-prompt representations for both OpenVLA base and OpenVLA-LIBERO90, but decoded action sensitivity remains low.}
\label{tab:app_hidden_probe_full}
\end{table}

\subsection{Pixel-Relation Hidden-State Probe}
\label{app:pixel_relation_hidden_probe}

We also probe relation-defined examples where the target name is absent from the instruction. The image-dependent subset contains examples where the full image+relation representation recovers the target but the language-only or shuffled-image control fails on the same test example. This subset checks that the result is not explained only by the text template or dataset priors.

\begin{table}[H]
\centering
\small
\setlength{\tabcolsep}{3pt}
\renewcommand{\arraystretch}{0.92}
\begin{adjustbox}{max width=\columnwidth}
\begin{tabular}{lrrrr}
\toprule
Subset & $N$ & Schema Sens. & Action Sens. & Gap \\
\midrule
Full all test pairs & 180 & 1.000 [1.000, 1.000] & 0.094 [0.056, 0.139] & 0.906 [0.861, 0.944] \\
Language-only & 180 & 0.594 [0.522, 0.667] & 0.094 [0.056, 0.139] & 0.500 [0.417, 0.583] \\
Shuffled-image & 180 & 0.656 [0.583, 0.722] & 0.094 [0.056, 0.139] & 0.561 [0.483, 0.639] \\
Image-dependent subset & 100 & 1.000 [1.000, 1.000] & 0.060 [0.020, 0.110] & 0.940 [0.890, 0.980] \\
\bottomrule
\end{tabular}
\end{adjustbox}
\caption{Pixel-relation hidden-state probe with bootstrap confidence intervals. Even on examples where image information is needed to recover the relation-defined target, the decoded action remains weakly sensitive.}
\label{tab:app_pixel_relation_hidden_probe_ci}
\end{table}

\FloatBarrier

\FloatBarrier
\subsection{Activation Patching}
\label{app:activation_patching}

We include activation patching only as a preliminary intervention check. The
patching experiment uses 100 target-swap examples, OpenVLA base, layer 16, and
$\alpha=1.0$, with the native OpenVLA action decoder unchanged. The limited
change in decoded action sensitivity means the hidden-state probe should be
interpreted as evidence of linear decodability with weak action transfer, not as
proof that the decoder causally ignores the representation or that a simple
hidden-state edit repairs low-level control.

\begin{table}[H]
\centering
\small
\begin{adjustbox}{max width=\columnwidth}
\begin{tabular}{lr}
\toprule
Activation-patching metric & Value \\
\midrule
Native action sensitivity & 0.080 \\
Patched action sensitivity & 0.070 \\
Patch moves toward CF action & 0.370 \\
Mean native $\Delta$(original, CF) & 1.290 \\
Mean patched $\Delta$(original, patched) & 1.135 \\
Mean distance patched-to-CF action & 1.344 \\
\bottomrule
\end{tabular}
\end{adjustbox}
\caption{
Preliminary layer-16 activation patching on 100 target-swap examples.
Patching counterfactual hidden states into the original prompt has limited
effect on decoded actions, supporting the interpretation that readable semantic
information is not straightforwardly transferred to the action decoder.
}
\label{tab:app_activation_patching}
\end{table}

\FloatBarrier

\subsection{OpenVLA QA-Style Target Probe Negative Result}

\begin{table}[H]
\centering
\small
\begin{adjustbox}{max width=\linewidth}
\begin{tabular}{lr}
\toprule
Metric & Value \\
\midrule
Examples completed & 20/20 \\
Qwen schema sensitivity & 1.000 \\
OpenVLA QA-style schema sensitivity & 0.000 \\
Original target correct & 0.500 \\
Counterfactual target correct & 0.050 \\
\bottomrule
\end{tabular}
\end{adjustbox}
\caption{
Negative diagnostic.
Direct QA-style probing of the action-tuned OpenVLA checkpoint is unreliable and is not used as evidence of VLA-side semantic understanding.
}
\label{tab:app_openvla_qa_probe}
\end{table}

\FloatBarrier
\subsection{Short-Horizon Target Approach}
\label{app:short_horizon_300}

We expand the short-horizon study to 300 paired
OpenVLA-LIBERO90 rollouts, corresponding to 600
actual rollouts. The evaluation covers six LIBERO
spatial tasks with 50 trials per task and a horizon of
$K=20$ policy steps. Each pair starts from the same
initial state and compares the original instruction
rollout with the counterfactual target-instruction
rollout.

\begin{table}[H]
\centering
\small
\setlength{\tabcolsep}{4pt}
\begin{adjustbox}{max width=\linewidth}
\begin{tabular}{lrrrr}
\toprule
Method & $N$ & CF-target approach & Original-target approach & Mean target pref. \\
\midrule
Original instruction & 300 & 0.797 & 0.677 & 0.000247 \\
Counterfactual instruction & 300 & 0.697 & 0.640 & 0.000166 \\
\bottomrule
\end{tabular}
\end{adjustbox}
\caption{
Raw paired rollout results over 300 paired $K=20$ rollouts.
Counterfactual instructions do not reliably redirect raw OpenVLA trajectories by themselves, so we do not interpret this as a task-success result.
}
\label{tab:app_short_horizon_raw_300}
\end{table}

\subsection{ActCheck Predicts Short-Horizon Target Preference}
\label{app:actcheck_predictive_rollout_300}

We stratify the same 300 paired $K=20$ rollouts by the
first-step ActCheck label. The results show that the
one-step target-consistency label is predictive of
trajectory-level behavior: aligned first actions lead to
higher counterfactual-target approach rates and higher
final and integrated target-preference scores.

\begin{table}[H]
\centering
\small
\setlength{\tabcolsep}{4pt}
\begin{adjustbox}{max width=\linewidth}
\begin{tabular}{lrrrr}
\toprule
First-step ActCheck label & $N$ & CF approach & Final target pref. & Integrated target pref. \\
\midrule
Aligned & 86 & 0.872 & 0.0660 & 0.0659 \\
Wrong-target & 63 & 0.571 & 0.0306 & 0.0312 \\
Ambiguous & 151 & 0.649 & -0.0095 & -0.0097 \\
\midrule
Overall & 300 & 0.697 & 0.0206 & 0.0206 \\
\bottomrule
\end{tabular}
\end{adjustbox}
\caption{
Trajectory outcomes stratified by first-step ActCheck label.
Target-aligned first actions lead to the strongest short-horizon preference for the counterfactual target.
}
\label{tab:app_actcheck_groups_300}
\end{table}

\begin{table}[H]
\centering
\small
\setlength{\tabcolsep}{3pt}
\begin{adjustbox}{max width=\linewidth}
\begin{tabular}{llccc}
\toprule
Metric & Comparison & Diff. & 95\% CI & Perm. $p$ \\
\midrule
Final target preference & aligned -- ambiguous & 0.0755 & [0.0679, 0.0829] & 0.0001 \\
Final target preference & aligned -- wrong-target & 0.0354 & [0.0272, 0.0431] & 0.0001 \\
Integrated target preference & aligned -- ambiguous & 0.0756 & [0.0679, 0.0830] & 0.0001 \\
Integrated target preference & aligned -- wrong-target & 0.0348 & [0.0266, 0.0425] & 0.0001 \\
CF approach rate & aligned -- ambiguous & 0.2231 & [0.1186, 0.3260] & 0.0005 \\
CF approach rate & aligned -- wrong-target & 0.3007 & [0.1589, 0.4393] & 0.0001 \\
\bottomrule
\end{tabular}
\end{adjustbox}
\caption{
Permutation-test comparisons for ActCheck-stratified $K=20$ rollouts.
Aligned first-step actions significantly outperform both ambiguous and wrong-target first-step actions on final target preference, integrated target preference, and counterfactual-target approach rate.
}
\label{tab:app_actcheck_significance_300}
\end{table}

\subsection{Schema-Conditioned Projection Stress Test}

The schema-conditioned projection study is a proof-of-concept action-interface intervention rather than a deployable controller. The target position is obtained from simulator-accessible object states. Given end-effector position $p_e$ and target position $p_t$, the translational action is projected toward the target direction:
\begin{equation}
    a'_{1:3}
    =
    \mathrm{dir}(p_t - p_e)
    \cdot
    \|a_{1:3}\|_2
    \cdot
    \lambda .
\end{equation}
This intervention tests whether explicit schema-level target information can redirect the action stream when injected at the action interface. It does not solve grasping, placing, contact dynamics, or long-horizon task completion.

\begin{table}[H]
\centering
\small
\begin{adjustbox}{max width=\linewidth}
\begin{tabular}{lrrrr}
\toprule
Intervention & $N$ & Raw Dist. $\downarrow$ & Projected Dist. $\downarrow$ & Full Success \\
\midrule
Always-project & 5 & 0.336 & 0.038 & 0/5 \\
Until-close & 5 & 0.336 & 0.065 & 0/5 \\
Phase-aware adapter & 3 & 0.332 & 0.090 & 0/3 \\
\bottomrule
\end{tabular}
\end{adjustbox}
\caption{
Closed-loop target-swap summary.
Schema-conditioned interventions redirect trajectories toward counterfactual schema targets, but full LIBERO task success remains zero.
}
\label{tab:app_projection_summary}
\end{table}

\begin{table}[H]
\centering
\small
\begin{adjustbox}{max width=\linewidth}
\begin{tabular}{lrrr}
\toprule
Setting & Release Rate & Close-but-Fail & Success \\
\midrule
d005 & 0.8 & 0.8 & 0 \\
d008 & 1.0 & 1.0 & 0 \\
d010 & 1.0 & 1.0 & 0 \\
\bottomrule
\end{tabular}
\end{adjustbox}
\caption{
Until-close projection.
Projection reliably brings the robot close to the schema target, but full success remains zero.
}
\label{tab:app_until_close}
\end{table}

\begin{table}[H]
\centering
\small
\begin{adjustbox}{max width=\linewidth}
\begin{tabular}{lrrrr}
\toprule
Horizon & Raw Final Dist. & Projection Final Dist. & Redirection & Success \\
\midrule
160 & 0.3434 & 0.0713 & +0.2721 & 0 \\
220 & 0.3357 & 0.0378 & +0.2978 & 0 \\
\bottomrule
\end{tabular}
\end{adjustbox}
\caption{
Horizon extension.
Longer rollouts bring the end-effector closer to the schema target, but full success remains zero.
}
\label{tab:app_horizon_extension}
\end{table}

\subsection{Phase-Aware Adapter}

\begin{table}[H]
\centering
\small
\begin{adjustbox}{max width=\linewidth}
\begin{tabular}{llrrrrr}
\toprule
Task & Target & Raw Dist. & Projected Dist. & Redirection & Lift $\Delta z$ & Lifted \\
\midrule
task1 & ramekin & 0.3890 & 0.0936 & +0.2954 & 0.1872 & yes \\
task3 & cookie box & 0.3799 & 0.0794 & +0.3005 & 0.0606 & yes \\
task5 & ramekin & 0.2264 & 0.0983 & +0.1281 & 0.0707 & yes \\
\bottomrule
\end{tabular}
\end{adjustbox}
\caption{
Phase-aware adapter results.
The adapter redirects the robot toward the schema target and lifts the target in all three tested counterfactual cases, although full pick-and-place success remains unsolved.
}
\label{tab:app_phase_adapter}
\end{table}

\begin{table}[H]
\centering
\small
\begin{adjustbox}{max width=\linewidth}
\begin{tabular}{lr}
\toprule
Metric & Value \\
\midrule
Mean raw final distance & 0.3318 \\
Mean projected final distance & 0.0904 \\
Mean redirection score & +0.2413 \\
Target lift rate & 3/3 \\
Full LIBERO success & 0/3 \\
\bottomrule
\end{tabular}
\end{adjustbox}
\caption{Phase-aware adapter summary.}
\label{tab:app_phase_adapter_summary}
\end{table}

\paragraph{Takeaway.}
The rollout redirection studies support a supplementary interface-level claim:
explicitly injecting schema-level target information can redirect the VLA action
stream. Always-on and until-close projection reduce final distance to the
counterfactual schema target, and a small phase-aware adapter lifts the selected
target in three tested cases. Full task success remains zero in these
counterfactual rollouts, so these results are evidence of trajectory-level
redirection and partial target interaction rather than a complete controller.

\FloatBarrier
\clearpage

\section{Verified Proposal Selection and Behavioral Evaluation}
\label{app:verified_selection}

\subsection{VISA-Rerank Exposure Quality}

VISA-Rerank uses the recovered schema to verify a finite set of candidate
actions before exposure. Candidates that fail the semantic consistency check
are not exposed; if no candidate is verified, the interface defers. This is an
execution-time selection mechanism and does not update the underlying VLA
parameters.

On 600 target-swap examples, VISA-Rerank retains 0.938 coverage and eliminates
wrong or ambiguous exposure under the target-consistency verifier.

\begin{table}[H]
\centering
\scriptsize
\setlength{\tabcolsep}{4pt}
\begin{tabular}{@{}lccc@{}}
\toprule
Method & Allow & Align. & Wrong \\
\midrule
Native & -- & 0.682 & 0.318 \\
VISA-Rerank & 0.938 & 1.000 & 0.000 \\
\bottomrule
\end{tabular}
\caption{
Target-swap exposure quality. Allow denotes non-deferred coverage. Alignment
and wrong/ambiguous rates are measured among exposed actions. Because target
consistency is used during verification, this table is interpreted as
verifier-based exposure quality rather than independent downstream success.
}
\label{tab:app_visa_rerank_target_swap}
\end{table}

The same proposal-and-verification interface extends to the richer semantic
splits.

\begin{table}[H]
\centering
\scriptsize
\setlength{\tabcolsep}{4pt}
\begin{tabular}{@{}lccc@{}}
\toprule
Split & Native & Allow & Align. \\
\midrule
Compositional & 0.513 & 0.890 & 1.000 \\
Temporal/procedural & 0.590 & 0.907 & 1.000 \\
\bottomrule
\end{tabular}
\caption{
VISA-Rerank exposure quality on richer semantic interventions. Native denotes
the native target-alignment rate, Allow denotes VISA-Rerank coverage, and
Align. denotes target consistency among allowed actions.
}
\label{tab:app_visa_rerank_richer}
\end{table}

\subsection{Independent Rollout Evaluation}
\label{app:visa_rerank_rollouts}

Because \actcheck{} participates in reranking, we evaluate selected actions with
trajectory metrics that are not used by the verifier. At $K=100$, VISA-Rerank
increases both final and integrated target preference and reduces wrong-object
approach.

\begin{table}[H]
\centering
\scriptsize
\setlength{\tabcolsep}{4pt}
\begin{tabular}{@{}lccc@{}}
\toprule
Method & Final & Int. & Wrong obj. \\
\midrule
Native & 0.0002 & 0.0178 & 0.720 \\
VISA-Rerank & 0.0083 & 0.0216 & 0.600 \\
\bottomrule
\end{tabular}
\caption{
Independent $K=100$ behavioral evaluation. Final and Int. denote final and
integrated target preference; Wrong obj. denotes wrong-object approach. These
trajectory metrics are not used by the reranking verifier.
}
\label{tab:app_visa_rerank_independent}
\end{table}

A separate $K=50$, $N=100$ evaluation shows the same qualitative trend.

\subsection{True Environment-Step Pilot}
\label{app:selective_correction_pilot}
We additionally evaluate a small true environment-step pilot with ten
manipulation trials. The selective correction acts only on high-confidence
wrong-target approach actions and releases control back to the native policy
near the target.

\begin{table}[H]
\centering
\scriptsize
\setlength{\tabcolsep}{3pt}
\begin{tabular}{@{}lccc@{}}
\toprule
Method & Contact & Lift & Wrong \\
\midrule
Native & 0.90 & 0.70 & 0 / 0 \\
Generic selective residual & 0.90 & 0.90 & 0 / 0 \\
VISA selective residual & 0.90 & 0.80 & 0 / 0 \\
Oracle selective residual & 1.00 & 0.90 & 0 / 0 \\
\bottomrule
\end{tabular}
\caption{
Preliminary environment-step manipulation pilot ($N=10$). Wrong reports wrong
contact / wrong lift. VISA corrects 2/3 native lift failures, but the generic
selective baseline is stronger; we therefore do not claim VISA-specific
controller superiority.
}
\label{tab:app_true_env_pilot}
\end{table}

These experiments support a limited interface-level claim: recovered semantics
can be used to regulate which action is exposed and can affect short-horizon
behavior, without establishing a repaired policy or general long-horizon
manipulation competence.

\FloatBarrier
\clearpage

\section{Prompt Templates}
\label{app:prompts}

\subsection{Schema Prompt}

The schema prompt asks the VLM to produce a structured JSON-like object with the following fields:
\begin{itemize}
    \item \texttt{target},
    \item \texttt{target\_exists},
    \item \texttt{phase},
    \item \texttt{allowed\_actions},
    \item \texttt{blocked\_actions},
    \item \texttt{next\_action},
    \item \texttt{execution\_decision}.
\end{itemize}
The prompt instructs the model not to invent missing targets, not to infer a task from a blank instruction, and not to execute prohibited actions.

\subsection{Pixel-Grounded Relation Prompt}

For pixel-grounded relation instructions, the prompt additionally asks the model to identify:
\begin{itemize}
    \item the anchor object,
    \item the spatial relation,
    \item the object satisfying the relation,
    \item whether the target exists,
    \item the admissible next action.
\end{itemize}
The target object name is not present in the instruction; the model must infer it from the image.

\subsection{Direct Qwen Gate Prompt}

The Direct Qwen Gate prompt asks the VLM to output exactly one of:
\begin{itemize}
    \item \texttt{allow},
    \item \texttt{ask},
    \item \texttt{hold},
    \item \texttt{abort},
    \item \texttt{target-not-found}.
\end{itemize}
Unlike VISA, this baseline does not expose target, target existence, allowed actions, blocked actions, or next-action schema fields.

\FloatBarrier
\section{Reproducibility Artifacts}
\label{app:artifacts}


The following scripts and output directories were used to construct SAT-Bench,
evaluate semantic-to-action transfer, run VISA diagnostics, and reproduce the
main and supplementary analyses:
\begin{itemize}
    \item Pixel-grounded relation schema outputs:
    \path{outputs/linguistic_blindness/qwen25vl7b_visual_relation_schema_600}
    \item Pixel-grounded relation analysis:
    \path{outputs/linguistic_blindness/pixel_grounded_relation_analysis}
    \item \actcheck{} script:
    \path{scripts/lb_visa_actcheck.py}
    \item Target-name ActCheck summary:
    \path{outputs/linguistic_blindness/visa_actcheck_exp2/actcheck_summary.csv}
    \item Visual-relation ActCheck summary:
    \path{outputs/linguistic_blindness/visual_relation_actcheck/actcheck_summary.csv}
    \item Octo visual-relation action sensitivity:
    \path{outputs/linguistic_blindness/octo_visual_relation_action_sensitivity_600}
    \item Octo relation analysis:
    \path{outputs/linguistic_blindness/octo_visual_relation_analysis}
    \item Qualitative ActCheck and CI packet:
    \path{outputs/linguistic_blindness/qualitative_ci_packet}
    \item Qualitative ActCheck candidate cases:
    \path{outputs/linguistic_blindness/qualitative_ci_packet/qualitative_actcheck_case_candidates.csv}
    \item Selected qualitative ActCheck cases:
    \path{outputs/linguistic_blindness/qualitative_ci_packet/recommended_qualitative_cases.csv}
    \item Bootstrap CI supplement:
    \path{outputs/linguistic_blindness/qualitative_ci_packet/bootstrap_ci_core_supplement.csv}
    \item Templated invalid mitigation:
    \path{outputs/linguistic_blindness/invalid_instruction_mitigation}
    \item Non-template invalid benchmark:
    \path{outputs/linguistic_blindness/hard_invalid_set_300/benchmark.jsonl}
    \item Direct Qwen / VISA / hybrid risk--utility report:
    \path{outputs/linguistic_blindness/risk_utility_direct_visa_hybrid/risk_utility_report.md}
    \item Hidden-state probe outputs:
    \path{outputs/linguistic_blindness/p0_oral_main_outputs/hidden_state_probe_main_table.csv}
    \item Short-horizon target approach:
    \path{outputs/linguistic_blindness/p0_oral_main_outputs/paired_rollout_k20_100_bootstrap_ci.csv}
    \item Schema-conditioned projection rollouts:
    \path{outputs/linguistic_blindness/schema_conditioned_projection}
\end{itemize}

\section{BridgeData V2 Second-Domain Diagnostic}
\label{app:bridge_v2}

We construct targeted BridgeData V2 real-robot diagnostics from the first public TFDS shard. The invalid-instruction subset contains 52 trajectories, 165 sampled frames, 495 invalid or prohibited instructions, and 165 normal instructions. We also construct a 165-example fixed-frame target/task-change diagnostic by replacing the original Bridge instruction with a different target or task instruction from the same shard. Because this shard does not provide reliable object metadata, the Bridge target/task split should be interpreted as an instruction-change diagnostic rather than a strict object-grounded target-swap benchmark.

\begin{table}[H]
\centering
\small
\begin{adjustbox}{max width=\linewidth}
\begin{tabular}{lrrrr}
\toprule
Dataset / policy & $N$ & Sem. change & Act. sens. $\uparrow$ & Gap \\
\midrule
BridgeData V2 / Octo & 165 & 1.000 & 0.661 & 0.339 \\
\bottomrule
\end{tabular}
\end{adjustbox}
\caption{BridgeData V2 target/task-change diagnostic. Octo is more sensitive than OpenVLA on LIBERO, but 33.9\% of instruction changes remain below the Bridge-specific paraphrase threshold.}
\label{tab:app_bridge_target_change}
\end{table}

\begin{table}[H]
\centering
\small
\setlength{\tabcolsep}{5pt}
\begin{adjustbox}{max width=\linewidth}
\begin{tabular}{lrrrr}
\toprule
Perturbation & $N$ & Octo action inhib. $\uparrow$ & Octo blind exec. $\downarrow$ & VISA safe def. $\uparrow$ \\
\midrule
Blank & 165 & 0.527 & 0.473 & 1.000 \\
Impossible & 165 & 0.576 & 0.424 & 1.000 \\
Negation & 165 & 0.133 & 0.867 & 1.000 \\
\midrule
Overall & 495 & 0.412 & 0.588 & 1.000 \\
\bottomrule
\end{tabular}
\end{adjustbox}
\caption{BridgeData V2 invalid/prohibited instruction results. Octo still exhibits substantial blind execution, especially under negation; VISA schema-gated execution consistently defers invalid commands.}
\label{tab:app_bridge_action_visa}
\end{table}

\begin{table}[H]
\centering
\small
\begin{adjustbox}{max width=\linewidth}
\begin{tabular}{lrr}
\toprule
Split & $N$ & Direct Qwen safe def. $\uparrow$ \\
\midrule
Blank & 165 & 1.000 \\
Impossible & 165 & 1.000 \\
Negation & 165 & 0.030 \\
\midrule
Invalid overall & 495 & 0.677 \\
Normal pass & 165 & 0.515 \\
\bottomrule
\end{tabular}
\end{adjustbox}
\caption{Direct Qwen Gate on the BridgeData V2 diagnostic subset. The direct gate handles blank and impossible commands but is brittle under negation.}
\label{tab:app_bridge_direct_qwen}
\end{table}

\FloatBarrier
\section{Additional Robustness Diagnostics}
\label{app:additional_robustness}

This section collects robustness diagnostics for SAT-Bench and VISA. The
diagnostics are organized around alternative explanations for the main
semantic--action gap result: a single threshold choice, grounding failure,
OpenVLA-only behavior, spatial overlap between targets, first-step-only
artifacts, all-stop utility loss, and surface-form brittleness under invalid or
prohibited commands. Together, these checks support the interpretation of
semantic-to-action transfer as a graded action-interface property rather than a
single-model failure.
\subsection{Threshold-Free Action Sensitivity}
\label{app:threshold_free}

\begin{table}[H]
\centering
\small
\setlength{\tabcolsep}{3pt}
\begin{adjustbox}{max width=\linewidth}
\begin{tabular}{lrrrr}
\toprule
Setting & $N$ & Mean target $\Delta$ & Mean control $\Delta$ & AUC \\
\midrule
OpenVLA target-name swap & 600 & 1.268 & 1.108 & 0.573 \\
OpenVLA pixel-relation swap & 600 & 1.454 & 1.237 & 0.583 \\
BridgeData V2 Octo target/task change & 165 & 1.912 & 0.220 & 0.929 \\
\bottomrule
\end{tabular}
\end{adjustbox}
\caption{
Threshold-free action sensitivity summary.
OpenVLA target-change and paraphrase-control deltas are only weakly separated, while the same metric strongly separates target/task changes in the BridgeData V2 Octo diagnostic.
}
\label{tab:app_threshold_free_auc}
\end{table}

\subsection{Schema--Action Quadrant Attribution}
\label{app:schema_action_quadrant}

\begin{table}[H]
\centering
\small
\begin{adjustbox}{max width=\linewidth}
\begin{tabular}{lrr}
\toprule
Case type & Count & Rate \\
\midrule
Schema correct + action consistent & 311 & 0.518 \\
Schema correct + action wrong/ambiguous & 264 & 0.440 \\
Schema wrong + action consistent & 16 & 0.027 \\
Schema wrong + action wrong/ambiguous & 9 & 0.015 \\
\bottomrule
\end{tabular}
\end{adjustbox}
\caption{
Schema--action quadrants for the OpenVLA pixel-grounded relation split.
The dominant failure mode is correct schema grounding with a wrong or ambiguous native action, which localizes many errors to the semantic-to-action interface.
}
\label{tab:app_schema_action_quadrant}
\end{table}

\subsection{Policy and Domain Matrix}
\label{app:policy_domain_matrix}

\begin{table}[H]
\centering
\small
\setlength{\tabcolsep}{3pt}
\begin{adjustbox}{max width=\linewidth}
\begin{tabular}{llrrrr}
\toprule
Policy & Setting & $N$ & Sem./oracle & Act. sens. & SAG \\
\midrule
OpenVLA base & LIBERO target-name & 600 & 1.000 & 0.068 & 0.932 \\
OpenVLA-LIBERO90 & LIBERO target-name & 600 & 1.000 & 0.057 & 0.943 \\
Octo-small-1.5 & LIBERO target-name & 600 & 1.000 & 0.427 & 0.573 \\
OpenVLA base & LIBERO pixel relation & 600 & 0.958 & 0.077 & 0.881 \\
Octo-small-1.5 & LIBERO pixel relation & 600 & 0.958 & 0.375 & 0.583 \\
Octo-small-1.5 & Bridge target/task change & 165 & 1.000 & 0.661 & 0.339 \\
\bottomrule
\end{tabular}
\end{adjustbox}
\caption{
Policy/domain matrix for semantic--action gaps.
The result supports a graded behavioral-axis interpretation: the gap is not equally severe everywhere, but it remains measurable across settings.
}
\label{tab:app_policy_domain_matrix}
\end{table}

\subsection{Normal-Command Utility Breakdown}
\label{app:normal_utility}

\begin{table}[H]
\centering
\small
\begin{adjustbox}{max width=\linewidth}
\begin{tabular}{lrrrr}
\toprule
Command type & $N$ & Normal pass & False block & Target valid \\
\midrule
Overall & 600 & 0.940 & 0.060 & 1.000 \\
Spatial relation & 468 & 1.000 & 0.000 & 1.000 \\
Drawer/cabinet & 132 & 0.727 & 0.273 & 1.000 \\
\bottomrule
\end{tabular}
\end{adjustbox}
\caption{
VISA normal-command utility breakdown.
VISA is not an all-stop gate: it preserves all spatial-relation normal commands, while false blocks are concentrated in drawer/cabinet commands.
}
\label{tab:app_normal_utility_breakdown}
\end{table}

\subsection{Schema-Conditioned Directional Proxy}
\label{app:repair_proxy}

\begin{table}[H]
\centering
\small
\begin{adjustbox}{max width=\linewidth}
\begin{tabular}{lrrrr}
\toprule
Method & $N$ & Target-aligned & Wrong/ambiguous & Intervention rate \\
\midrule
Native OpenVLA action & 600 & 0.545 & 0.455 & 0.000 \\
\actcheck{} directional candidate & 600 & 1.000 & 0.000 & 0.455 \\
\bottomrule
\end{tabular}
\end{adjustbox}
\caption{
Schema-conditioned directional proxy.
The directional candidate is a post-hoc target-consistency diagnostic rather than
a deployable controller. It shows that recovered schema targets can define a
target-consistent directional alternative without establishing full control.
}
\label{tab:app_repair_proxy}
\end{table}

\subsection{Negation and Prohibition Stress Test}
\label{app:negation_stress}

The negation/prohibition stress split is constructed from 200 normal base examples and four variants per base example: \texttt{do\_not}, \texttt{avoid}, \texttt{leave\_alone}, and \texttt{except}, for 800 total prohibited commands.

\begin{table}[H]
\centering
\small
\begin{adjustbox}{max width=\linewidth}
\begin{tabular}{lrrr}
\toprule
Method & $N$ & Safe/Inhibited $\uparrow$ & Blind/Allow $\downarrow$ \\
\midrule
Octo action inhibition & 800 & 0.298 & 0.703 \\
Direct Qwen Gate & 800 & 0.365 & 0.635 \\
VISA / Qwen Schema Gate & 800 & 1.000 & 0.000 \\
\bottomrule
\end{tabular}
\end{adjustbox}
\caption{
Overall negation/prohibition stress results.
Octo continues executing most prohibited commands; direct binary gating allows most commands; the structured schema gate defers all prohibited variants in this controlled diagnostic split.
}
\label{tab:app_negation_overall}
\end{table}

\begin{table}[H]
\centering
\small
\begin{adjustbox}{max width=\linewidth}
\begin{tabular}{lrrr}
\toprule
Variant & Octo inhibited $\uparrow$ & Direct safe deferral $\uparrow$ & VISA safe deferral $\uparrow$ \\
\midrule
\texttt{avoid} & 0.205 & 1.000 & 1.000 \\
\texttt{do\_not} & 0.150 & 0.000 & 1.000 \\
\texttt{except} & 0.150 & 0.460 & 1.000 \\
\texttt{leave\_alone} & 0.685 & 0.000 & 1.000 \\
\bottomrule
\end{tabular}
\end{adjustbox}
\caption{
Variant-level negation/prohibition stress results.
Direct Qwen Gate is highly sensitive to surface form: it succeeds on \texttt{avoid}, partly handles \texttt{except}, and fails on \texttt{do\_not} and \texttt{leave\_alone}.
}
\label{tab:app_negation_by_variant}
\end{table}

\paragraph{Interpretation.}
These diagnostics make the main claim falsifiable rather than merely adding
additional benchmark numbers. Threshold-free AUCs test whether the result is a
single-cutoff artifact. Spatial subsets test whether low action sensitivity is
caused by overlapping target directions. Schema--action quadrants separate
grounding failures from semantic-to-action transfer failures. The policy/domain
matrix shows that the gap is graded: more responsive policies reduce the gap
but do not remove the need to report it. Normal-command utility prevents a
trivial all-stop gate from appearing safe, and the non-template invalid and
negation/prohibition splits test whether non-executable semantics inhibit
motion beyond surface-form matching. Taken together, these results support the
paper's interface-level interpretation: semantic recovery, native action
sensitivity, target consistency, and non-execution are distinct properties and
should be measured separately.

\end{document}